\documentclass[]{fairmeta}
\usepackage{amsmath,amssymb}
\usepackage{graphicx}
\usepackage[numbers]{natbib}
\usepackage{booktabs}
\usepackage{bm}
\usepackage{xspace}
\usepackage{tabularx}
\usepackage{multirow}
\usepackage{fontspec}
\usepackage{sectsty}
\usepackage{subcaption}
\usepackage{makecell}
\usepackage{array}
\usepackage{longtable}
\usepackage{needspace}
\usepackage[hidelinks,breaklinks]{hyperref}

\newfontfamily\calmregular[AutoFakeBold=2]{Outfit-Regular.ttf}
\newcommand{\calmfont}[1]{{{\calmregular #1}}}

\newcommand{\modelname}{HoloBrain-0.5\xspace}
\definecolor{scoregreen}{HTML}{187544}
\newcommand{\scoregain}[1]{\nobreak\hspace{0.35em}\textcolor{scoregreen}{+#1}}

\allsectionsfont{\calmfont}

\title{\calmfont{\modelname Technical Report}}
\author{\calmfont{Xuewu~Lin}}
\author{\calmfont{Jiahao~Wang}}
\author{\calmfont{Jiawei~Li}}
\author{\calmfont{Mengdi~Li}}
\author{\calmfont{Zhengmao~Sun}}
\author{\calmfont{Yiwei~Jin}}
\author{\calmfont{Hongyu~Xie}}
\author{\calmfont{Zhengyang~Feng}}
\author{\calmfont{Kun~Wu}}
\author{\calmfont{Zihao~Chen}}
\author{\calmfont{Lianzhou~Zhang}}
\author{\calmfont{Lichao~Huang}}
\author[*]{\calmfont{Zhizhong~Su}}
\affiliation{\calmfont{Horizon Robotics HoloBrain Team}}
\contribution[*]{Corresponding author.}
\code{\href{https://github.com/HorizonRobotics/HoloBrain}{HoloBrain}\quad\textbar\quad\href{https://github.com/HorizonRobotics/UniURDF}{UniURDF}}
\metadata[Model]{\href{https://huggingface.co/HorizonRobotics/HoloBrain_v0.5}{Model weights on Hugging Face}}
\webpage{\url{horizonrobotics.github.io/robot_lab/holobrain_0_5}}
\correspondence{Zhizhong Su at \email{zhizhong.su@horizon.auto}}

\begin{document}

\abstract{
We present HoloBrain-0.5, a family of vision-language-action (VLA) models for manipulation across heterogeneous embodiments.
Our goal is to plan motion in a shared action space across embodiments, using the robot state, kinematic structure, and calibrated camera observations as inputs.
Building on HoloBrain-0, we introduce three improvements.
First, we standardize joint axes, coordinate frames, and tool-center points in each robot description (URDF), then encode its kinematic graph in the policy.
Second, we release HoloBrain-0.5-Continuous, a diffusion policy, and HoloBrain-0.5-Discrete, which quantizes actions with residual finite scalar quantization (FSQ) and generates the resulting codes by delayed autoregression. Both use the same 3D embodiment representation, which accommodates arbitrary kinematic configurations.
Third, we scale pretraining to more than 10,000 hours of real-robot, simulation, and human-manipulation data spanning 15 embodiment categories, with calibrated camera geometry and canonicalized embodiment representations.

During pretraining, the discrete policy converges faster than the continuous policy and achieves a higher final mean success rate across six closed-loop simulation settings (56.77\% versus 49.71\%).
On seven real-robot tasks, the discrete policy achieves a mean progress score of 87.76\% without task-specific post-training. After post-training, the continuous policy outperforms $\pi_{0.5}$ on average on both the seven-task benchmark (94.52\% versus 86.25\%) and an 11-task dual-arm benchmark (79.07\% versus 69.41\%).
The discrete-initialized continuous policy also transfers five tasks zero-shot from Piper to Piper-X, scoring 63.28\% on average without demonstrations of those tasks on Piper-X.
With target-robot demonstrations, the continuous policy reaches mean progress scores of 89.2\% on a single-arm Franka Panda and 92.94\% on a mobile dual-arm platform. Although pretraining includes no Unitree G1 data, post-training with a limited set of G1 demonstrations yields a mean progress score of 70.24\% across three humanoid tasks, compared with 11.00\% for $\pi_{0.5}$.
}

\maketitle

\section{Introduction}
\label{intro}

Vision-language-action (VLA) models ground visual observations and language instructions in executable robot actions, providing a scalable route toward general-purpose manipulation. Early large-scale systems established the value of vision-language pretraining and heterogeneous robot data~\cite{rt1,rt2,openx}, while recent models have improved action generation, open-world generalization, and cross-embodiment learning~\cite{pi0,pi05,pi07,walloss05,g05,lingbotvla}.

\begin{figure*}[!htbp]
    \centering
    \includegraphics[width=0.99\linewidth]{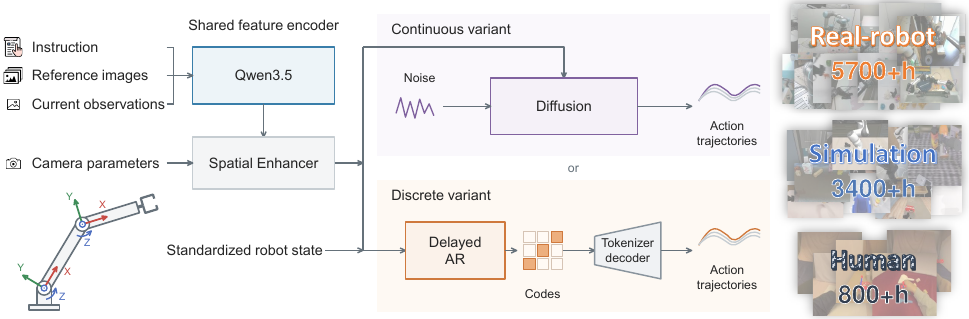}
    \caption{Overview of HoloBrain-0.5. Shared multimodal features and standardized robot state condition either the continuous diffusion policy or the discrete delayed-autoregressive policy with residual-FSQ tokenizer decoding. Both produce action trajectories. Standardized robot state includes embodiment geometry. The illustrated joint frames use $+x$ along the child link and outward $+z$ as the rotation axis; circled dots denote axes pointing out of the page. Pretraining combines real-robot, simulation, and human-manipulation data, illustrated at right. Robot geometry, noise, trajectories, and the code grid are schematic.}
    \label{fig:overview}
\end{figure*}

A central challenge is that robot datasets differ not only in visual appearance and task distribution, but also in joint count, kinematic structure, camera configuration, and action conventions. Arbitrary coordinate choices in robot descriptions further obscure the relationship between an action value and its physical effect. HoloBrain-0 addressed part of this problem with a joint-centric architecture and explicit forward-kinematics features~\cite{holobrain0}. HoloBrain-0.5 extends this design with a canonical embodiment representation and two complementary action formulations. An overview is shown in Figure~\ref{fig:overview}.

We introduce URDF standardization to remove embodiment-specific coordinate conventions before policy learning. Revolute joints are reparameterized about their local $z$-axes, prismatic joints translate along the positive local $z$-axis, and tool-center-point (TCP) frames point outward along $z$.

HoloBrain-0.5 provides two model variants under a largely shared architecture. HoloBrain-0.5-Continuous models continuous action chunks with diffusion, whereas HoloBrain-0.5-Discrete applies finite scalar quantization (FSQ)~\cite{fsq} in successive residual stages and multidimensional delayed autoregressive generation. The discrete formulation exposes an explicit token distribution and converges faster in our pretraining experiments. We pretrain both variants on more than 10,000 hours of data across 15 embodiment categories.

We track closed-loop performance at intermediate pretraining checkpoints using six simulation settings from three simulators, with a total of 21,430 evaluation episodes per model checkpoint across the six settings. We additionally evaluate out-of-the-box behavior, real-robot post-training, zero-shot task transfer between seen embodiments, and adaptation across substantially different robot morphologies. The first two real-robot benchmarks contain 18 tasks in total, while the additional morphology-adaptation suite spans 11 tasks across three further embodiments.

Our main contributions are summarized as follows:
\begin{itemize}
    \item We introduce a canonical 3D embodiment representation and a shared action space based on standardized URDFs. Both are defined on the kinematic graph rather than a fixed set of joints.
    \item We release two policies built on this representation: the discrete policy converges faster during pretraining, while the continuous policy performs better after post-training.
    \item In real-robot trials, HoloBrain-0.5 improves on HoloBrain-0 in closed-loop performance and performs complex dexterous tasks. It achieves 63.28\% mean progress on tasks transferred zero-shot to a target robot and, with target-robot demonstrations, adapts to single-arm, humanoid, and mobile platforms.
\end{itemize}

\section{Related Work}

\subsection{Vision-Language-Action Models}

RT-1 and RT-2 established scalable vision-language-conditioned robot control and transfer from web-scale representations~\cite{rt1,rt2}. Open X-Embodiment~\cite{openx}, Octo~\cite{octo}, and OpenVLA~\cite{openvla} subsequently demonstrated the value of heterogeneous multi-robot pretraining. Recent systems pair large vision-language backbones with dedicated action modules: RDT-1B~\cite{rdt1b}, $\pi_0$~\cite{pi0}, CogACT~\cite{cogact}, and GR00T N1~\cite{grootn1} use diffusion- or flow-based generation, while $\pi_{0.5}$~\cite{pi05} and $\pi_{0.7}$~\cite{pi07} emphasize heterogeneous co-training and contextual conditioning.

Cross-embodiment scaling has also been explored through embodiment prompts or learned action representations in X-VLA~\cite{xvla}, LingBot-VLA~\cite{lingbotvla}, Wall-OSS-0.5~\cite{walloss05}, and G0.5~\cite{g05}. In parallel, RoboMamba~\cite{robomamba}, TinyVLA~\cite{tinyvla}, NORA~\cite{nora}, SmolVLA~\cite{smolvla}, and VLA-Adapter~\cite{vlaadapter} study more efficient VLA architectures. CEI~\cite{cei} transfers demonstrations by aligning trajectories across functionally similar end effectors in 3D. HoloBrain-0.5 instead standardizes joint axes, coordinate frames, and end-effector conventions so that demonstrations from different robots can share an action representation.

\subsection{Action-Space Modeling}

Action-space design determines how a policy represents temporal structure and multimodality. Continuous policies use direct action chunks~\cite{act} or generative objectives such as diffusion and flow matching~\cite{diffusionpolicy,rdt1b,pi0,cogact,grootn1}. Discrete policies instead quantize actions and predict token sequences. OpenVLA~\cite{openvla} uses discretized action tokens, FAST~\cite{fast} compresses action chunks in the frequency domain, and SpatialVLA~\cite{spatialvla} and G0.5~\cite{g05} introduce spatially or embodiment-aware tokenizations. FSQ~\cite{fsq} offers a codebook-free mechanism for constructing compact discrete representations.
VQ-BeT~\cite{vqbet} learns a hierarchy of discrete action codes through residual vector quantization.

Both HoloBrain-0.5 variants use the same canonical 3D joint graph to represent the embodiment and express actions as temporal sequences of joint graphs. This graph provides a complete kinematic representation of robot state and action, grounding both in the same 3D frame as calibrated visual observations. The continuous variant models these sequences with diffusion. The discrete variant uses residual FSQ to learn a discrete representation of the action prior from trajectories, then predicts the resulting codes by delayed autoregression.

\section{Method}

HoloBrain-0.5 retains the joint-centric policy interface of HoloBrain-0 while revising both the embodiment representation and action decoder. At control step $h$, the policy takes a language instruction $\boldsymbol{\tau}$, synchronized camera observations $\mathcal{I}_h=\{I_h^{(m)}\}_{m=1}^{M}$, the numerical robot state $\mathbf{s}_h$, the standardized URDF $\mathcal{U}$, and camera calibration parameters $\mathcal{C}=\{K_m,T_{B C_m}\}_{m=1}^{M}$, where $K_m$ and $T_{B C_m}$ are the intrinsic matrix and robot-base extrinsic transform of camera $m$, respectively. It predicts a future action chunk $\mathbf{A}_{h:h+H}$ according to
\begin{equation}
    \mathbf{A}_{h:h+H}
    \sim \pi_{\theta}\!\left(\cdot\mid
    \mathcal{I}_h,\boldsymbol{\tau},\mathbf{s}_h,\mathcal{U},\mathcal{C}\right).
    \label{eq:problem-formulation}
\end{equation}
Unlike a conventional VLA interface, HoloBrain explicitly supplies the embodiment description and calibrated camera geometry. The URDF is converted into the canonical joint graph used for action modeling, while camera calibration aligns visual observations with the robot-base frame. Section~3.1 describes this standardization, followed by the continuous and discrete decoders and the pretraining and calibration pipeline in Sections~3.2--3.4.

\subsection{URDF Standardization}

\begin{figure}[!htbp]
    \centering
    \includegraphics[width=\linewidth]{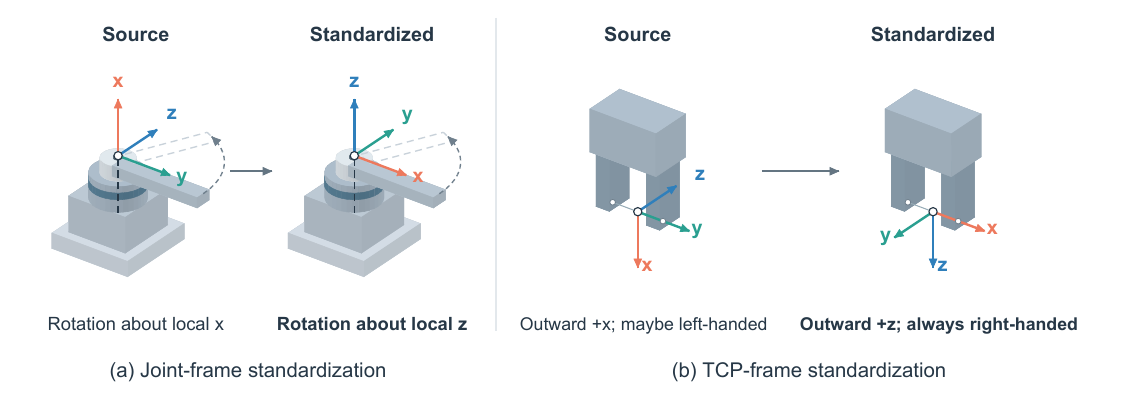}
    \caption{Joint and TCP frame standardization. (a) The joint's right-handed source frame is reoriented from local $x$-axis motion to local $z$-axis motion. Attached-transform compensation preserves physical geometry and joint trajectories; dashed geometry shows a second configuration. (b) A schematic left-handed dataset TCP frame is converted to a right-handed frame with outward $+z$. Handedness conversion is separate from the proper rotation in (a). The TCP origin is the midpoint of the two marked fingertip centers. Axis colors denote $x$ (red), $y$ (green), and $z$ (blue). Source conventions and geometry are illustrative; wrist roll is set per embodiment.}
    \label{fig:urdf-standardization}
\end{figure}

HoloBrain constructs its embodiment representation from a pruned URDF: it computes forward-kinematics (FK) losses over the retained kinematic tree, uses the 6-DoF poses of selected links to construct the joint-centric action space, and embeds the graph-hop distances among these links~\cite{holobrain0}. Consequently, URDF frame conventions directly enter the learned representation. Source URDFs may assign different local axes to physically equivalent joints, while manually selected representative links may be offset from the corresponding joint origins. We standardize each embodiment using a common, joint-centered convention, as illustrated in Figure~\ref{fig:urdf-standardization}.

We canonicalize every actuated joint and use the direct child link of each selected joint as its pose anchor. For joint $j$, let $P_j$ and $C_j$ denote its parent and direct child links, $q_j$ its joint value, and $\mathbf a_j$ its original unit motion axis. We write $T_{AB}$ for the transform from frame $B$ to frame $A$, with $T_{o,j}=T_{P_jC_j}(0)\in SE(3)$ denoting the joint origin. We choose a proper rotation $S_j\in SO(3)$ satisfying $S_j\hat{\mathbf z}=\mathbf a_j$, where $\hat{\mathbf z}=[0,0,1]^\top$. We then post-rotate the joint origin by $S_j$ and replace its axis with $\hat{\mathbf z}$; this preserves the origin translation while expressing revolute, continuous, and prismatic motion about or along the positive local $z$-axis. Axis alignment leaves the roll around $z$ unconstrained, so a manually specified \texttt{rot\_z} selects a consistent wrist-frame roll.

\begin{figure}[!htbp]
    \centering
    \begin{subfigure}[t]{0.49\linewidth}
        \centering\includegraphics[width=\linewidth]{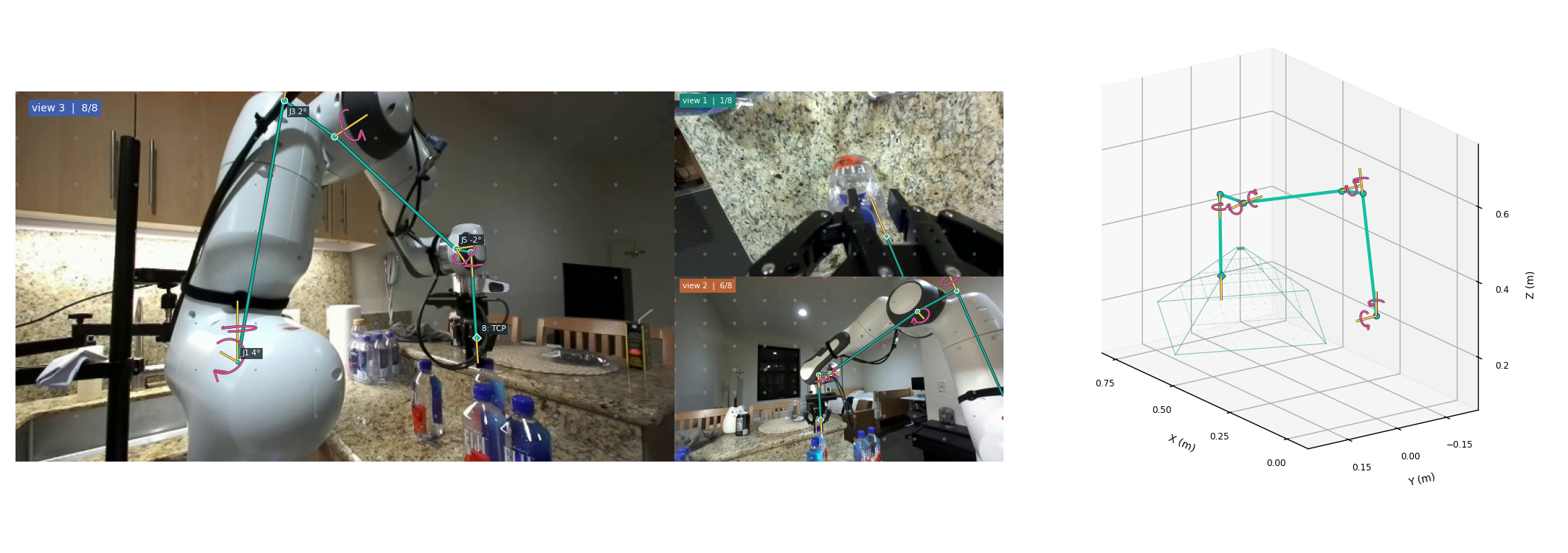}
        \caption{DROID Franka Panda (8 nodes)}
    \end{subfigure}\hfill
    \begin{subfigure}[t]{0.49\linewidth}
        \centering\includegraphics[width=\linewidth]{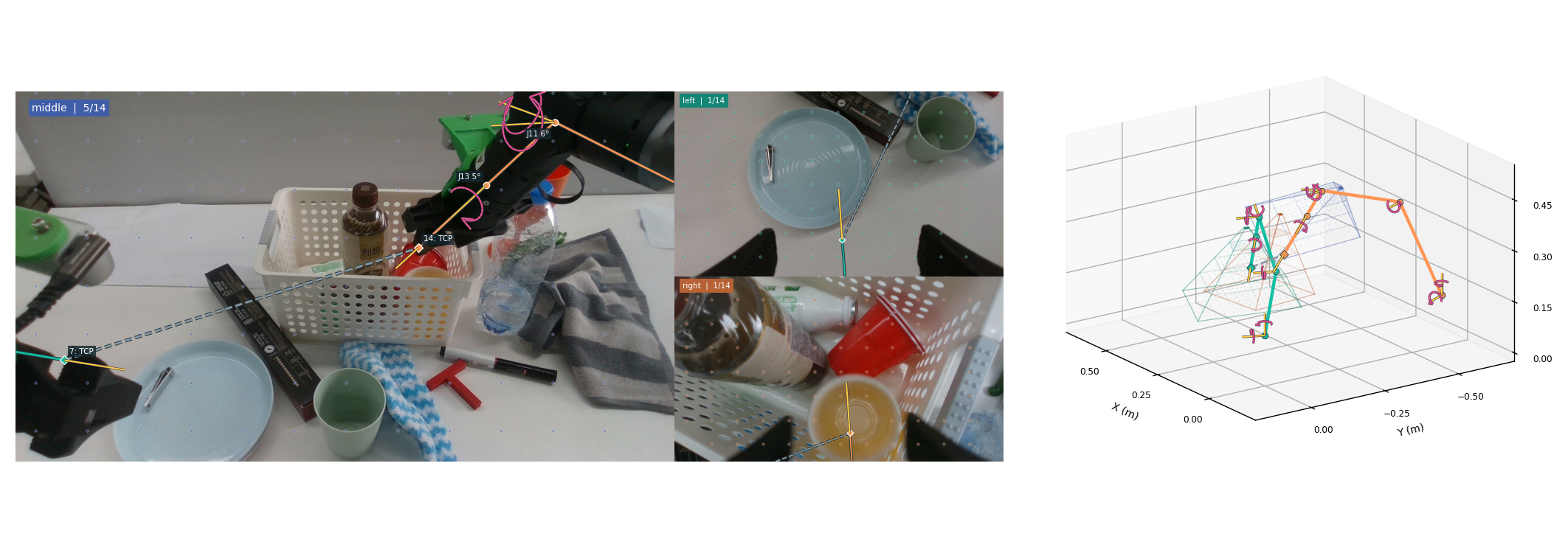}
        \caption{AgileX Piper (14 nodes)}
    \end{subfigure}
    \par\medskip
\begin{subfigure}[t]{0.49\linewidth}
        \centering\includegraphics[width=\linewidth]{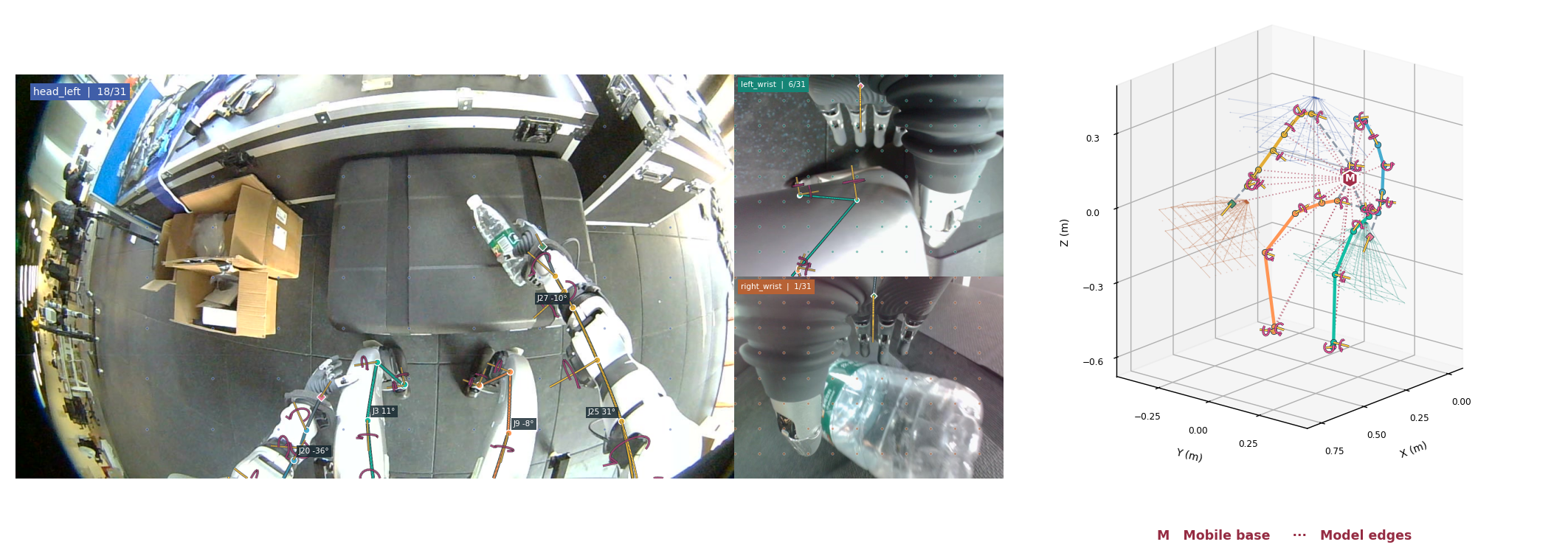}
        \caption{Unitree G1 (31 nodes)}
    \end{subfigure}\hfill
    \begin{subfigure}[t]{0.49\linewidth}
        \centering\includegraphics[width=\linewidth]{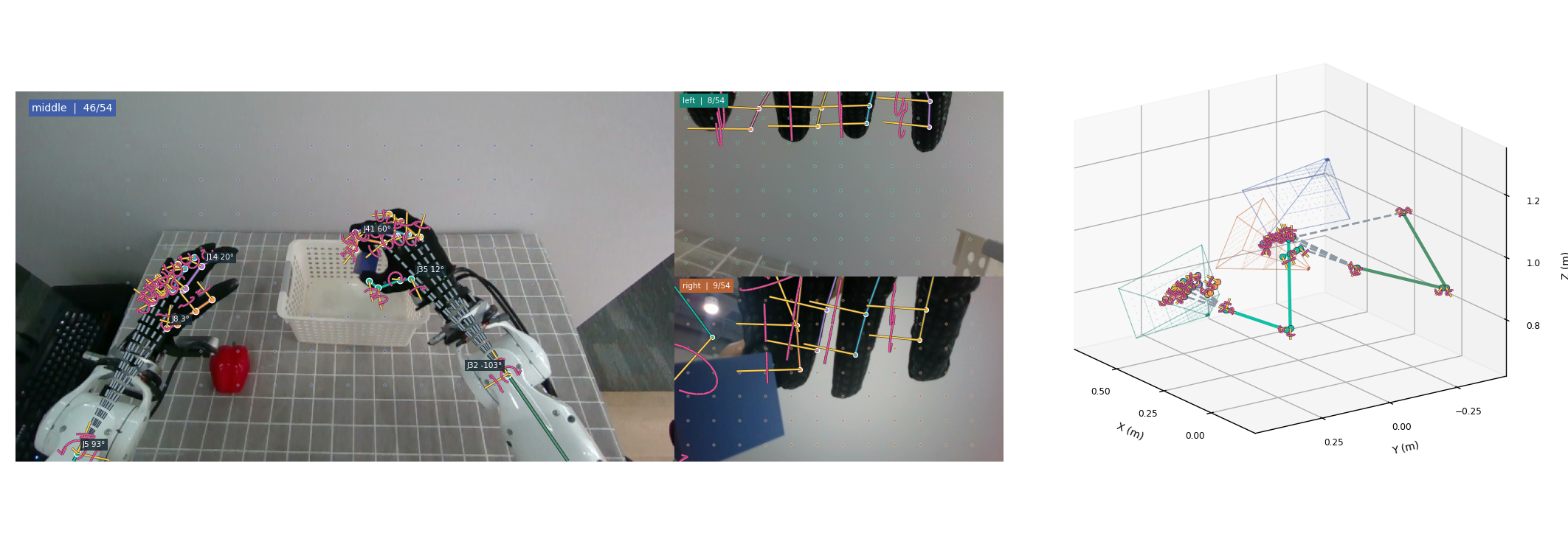}
        \caption{Tianji WuJi (54 nodes)}
    \end{subfigure}
    \caption{Canonical joint graphs on four robots. Each panel shows the physical nodes projected into the available camera views (left) and expressed in the robot-base frame (right).}
    \label{fig:joint-graph-examples}
\end{figure}

Figure~\ref{fig:joint-graph-examples} shows how the same action representation applies to a single arm, two dual-arm robots, and a humanoid. Each panel pairs camera projections with the same graph in the robot-base frame: node counts and camera layouts change, but a joint pose has the same geometric meaning across views and embodiments. Appendix~\ref{app:joint-graph-examples} gives further examples, including mobile-base and TCP-only interfaces.

\Needspace{14\baselineskip}
This frame rewrite changes the model-facing orientation of $C_j$ but not the robot's physical behavior. Let $\widetilde S_j\in SE(3)$ be the homogeneous lift of $S_j$, and let $\mathcal M(\mathbf u,q)$ denote revolute or prismatic motion about or along axis $\mathbf u$. The rewrite and compensation are
\begin{equation}
    \begin{aligned}
        \mathcal M(\mathbf a_j,q_j)
                 & =\widetilde S_j\mathcal M(\hat{\mathbf z},q_j)\widetilde S_j^{-1},                                 \\
        T'_{o,j} & =T_{o,j}\widetilde S_j,
                 & T'_{C_jX}                                                          & =\widetilde S_j^{-1}T_{C_jX},
    \end{aligned}
    \label{eq:urdf-standardization}
\end{equation}
where $X$ denotes a child-local visual, collision, inertial, or outgoing-joint frame. It follows that $T'_{P_jC_j}(q_j)=T_{P_jC_j}(q_j)\widetilde S_j$, whereas
\begin{equation}
    T'_{P_jX}(q_j)
    =T'_{P_jC_j}(q_j)T'_{C_jX}
    =T_{P_jC_j}(q_j)T_{C_jX}
    =T_{P_jX}(q_j).
    \label{eq:urdf-invariance}
\end{equation}
Thus, the child-link frame acquires the desired canonical orientation, but attached geometry, inertial frames, and downstream kinematics retain their original poses for every $q_j$. Recorded joint trajectories therefore remain valid. If a camera extrinsic references a reoriented link, we add a fixed compatibility frame that recovers the original reference frame.

We complete the representation by standardizing end-effector frames and pruning to the model-relevant graph. Each embodiment defines an outward-facing TCP $+z$ and a wrist-roll \texttt{rot\_z} once for all trajectories. We retain selected joints' child links, the declared TCP, gripper, and camera-reference links, and their root paths; other branches are deleted without changing retained transforms. HoloBrain evaluates FK on this pruned URDF, uses the selected child-link poses as its 6-DoF action representation, and embeds their hop distances.

We apply this procedure to all 15 embodiments. For each standardized URDF, we compare shared-link positions and visual-geometry poses with the source description at the zero configuration and at two deterministic nonzero configurations, ensuring that the check covers joint motion rather than only the rest pose. This process removes arbitrary frame conventions while preserving the physical semantics of the original trajectories.

\subsection{HoloBrain-0.5-Continuous}

The continuous variant follows the joint-centric encoder-decoder design and diffusion objective of HoloBrain-0~\cite{holobrain0}. Unless stated otherwise, the input representation, forward-kinematics conditioning, action parameterization, diffusion training objective, and inference procedure remain unchanged. This section focuses on the architectural modifications introduced in HoloBrain-0.5.

\textbf{Backbone and model capacity.} We replace the Qwen2.5-VL-3B backbone~\cite{qwen25vl} with Qwen3.5-2B~\cite{qwen35} and increase the capacity of the action decoder. Specifically, the number of decoder layers is increased from 6 to 10, while the embedding dimension is increased from 256 to 384.

\begin{figure*}[!htbp]
    \centering
    \includegraphics[width=0.8\linewidth]{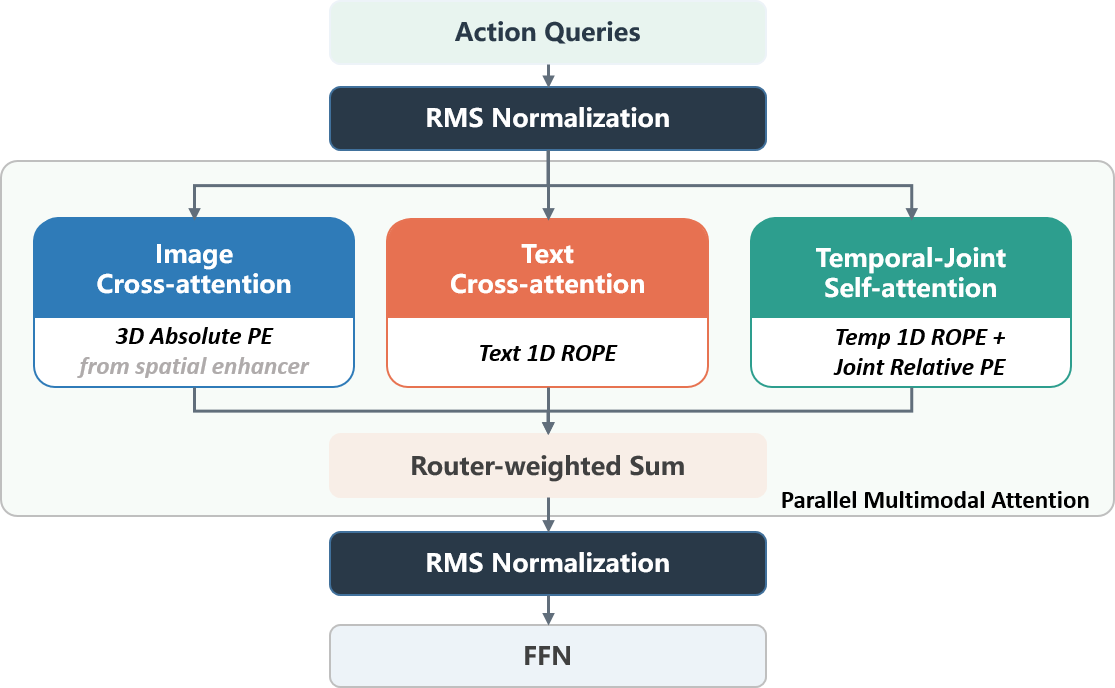}
    \caption{Parallel multimodal attention in the HoloBrain-0.5 action decoder. Image, text, and temporal-joint features are processed by independently parameterized attention branches and combined by a query- and head-dependent router through weighted summation.}
    \label{fig:attention}
\end{figure*}

\textbf{Parallel multimodal attention.} HoloBrain-0 applies image cross-attention, text cross-attention, and temporal-joint self-attention sequentially. We find that this ordering can bias the decoder toward the last processed modality. In particular, placing temporal-joint self-attention last encourages the policy to over-rely on proprioception and become less responsive to visual observations. HoloBrain-0.5 therefore computes the three attention branches in parallel and combines them with a query-dependent routing mechanism, as shown in Figure~\ref{fig:attention}.

This parallel design preserves modality-specific positional structures rather than imposing a shared positional representation across all branches.

Let $\mathbf{x}$ denote the action-query features, and let $\mathbf{C}_{v}$ and $\mathbf{C}_{t}$ denote the image and language features, respectively. We use three independently parameterized attention modules. Their outputs can be summarized as
\begin{equation}
\begin{aligned}
\mathbf{x}_{v} &= \operatorname{Attn}_{v}(\mathbf{x},\mathbf{C}_{v}^{3\mathrm{D}}),\\
\mathbf{x}_{t} &= \operatorname{Attn}_{t}\!\left(\mathbf{x},\operatorname{RoPE}_{1\mathrm{D}}(\mathbf{C}_{t})\right),\\
\mathbf{x}_{s} &= \operatorname{Attn}_{s}\!\left(\operatorname{JointRelative}(\operatorname{RoPE}(\mathbf{x}))\right),
\end{aligned}
\label{eq:parallel_attention}
\end{equation}
Here, $\mathbf{C}_{v}^{3\mathrm{D}}$ denotes image features equipped with 3D absolute positional encoding, $\operatorname{RoPE}_{1\mathrm{D}}$ denotes one-dimensional rotary positional encoding~\cite{roformer} for language features, and $\operatorname{JointRelative}(\operatorname{RoPE}(\mathbf{x}))$ denotes temporal rotary positional encoding combined with joint-relative positional encoding for the temporal-joint branch. The notation suppresses the query/key/value projections and residual connections inside each attention module.

The three branch outputs are combined by a query-dependent router. The router predicts a weight for each modality, and the final output is their weighted sum:
\begin{equation}
\boldsymbol{\alpha}=\operatorname{softmax}\!\left(\frac{\operatorname{Router}(\mathbf{x})}{\tau}\right),
\qquad
\mathbf{x}_{\mathrm{out}}
=\alpha_{v}\mathbf{x}_{v}+\alpha_{t}\mathbf{x}_{t}+\alpha_{s}\mathbf{x}_{s},
\label{eq:multimodal_router}
\end{equation}
where $\boldsymbol{\alpha}=(\alpha_v,\alpha_t,\alpha_s)$ and $\tau$ is the routing temperature. In the implementation, the routing weights are predicted independently for different attention heads; the head index is omitted here for readability.

\begin{figure*}[!htbp]
    \centering
    \includegraphics[width=0.99\linewidth]{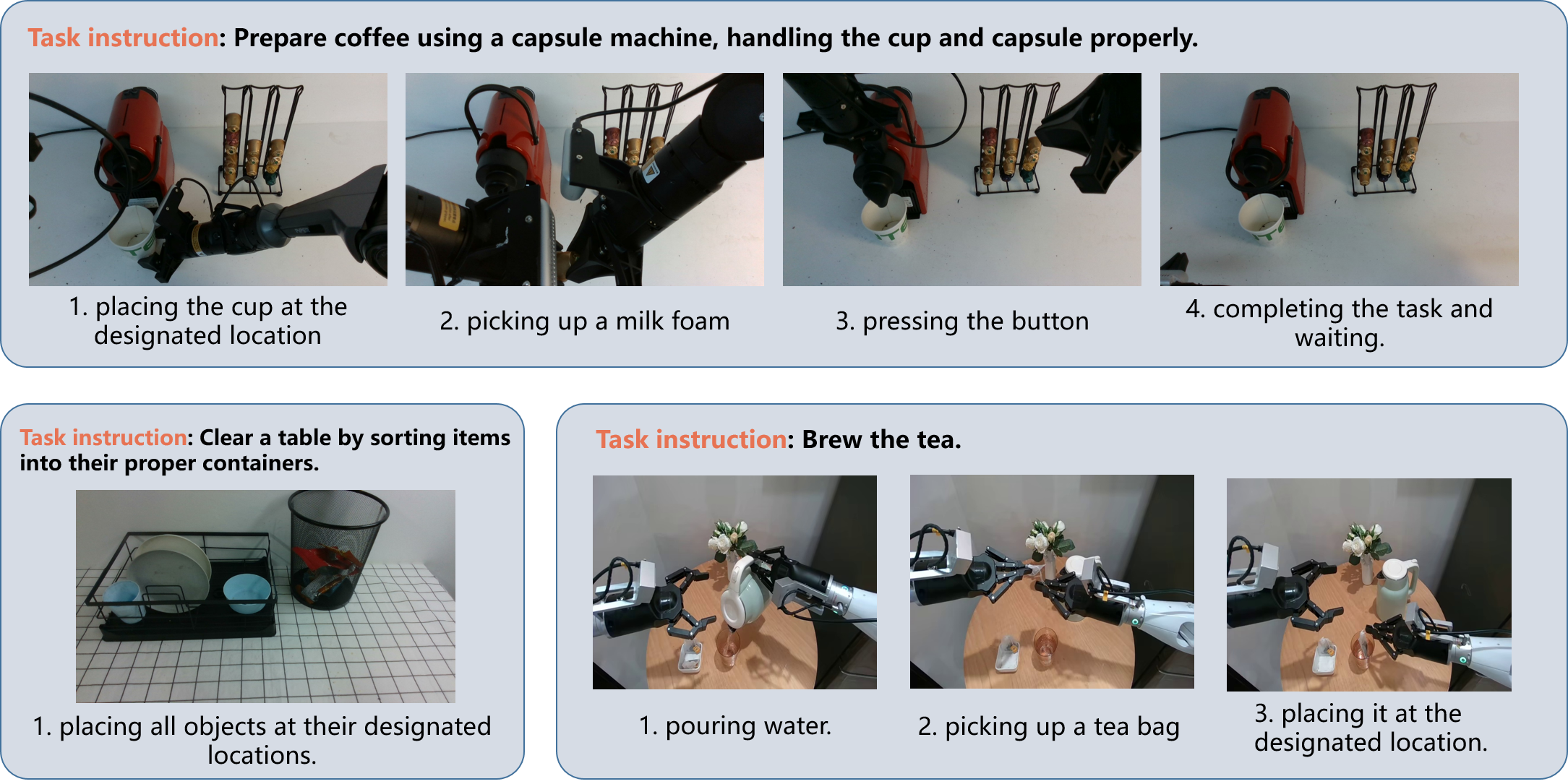}
    \caption{Reference images sampled from task-level key frames. (a, top) Preparing coffee: place the cup at the designated location, retrieve the milk-foam capsule, press the button, and wait for completion. (b, bottom left) Place all objects at their designated locations and finish. (c, bottom right) Brew tea: pour water, pick up a tea bag, and place it in the cup.}
    \label{fig:reference_images}
\end{figure*}

\textbf{Reference images from task-level key frames.} During pretraining, we randomly provide reference images for a subset of training samples. In many tasks, the textual instructions alone do not fully specify the task objective, the relevant scene configuration, or the desired execution process. As a result, the model may receive insufficient conditioning information to learn the intended visual-to-action mapping.
To provide more informative task conditions and support in-context learning, we annotate task-level key frames for a subset of tasks. During pretraining, a list of key frames is randomly sampled and the corresponding images are provided as reference images. Examples are shown in Figure~\ref{fig:reference_images}.

\begin{figure*}[!htbp]
    \centering
    \includegraphics[width=0.99\linewidth]{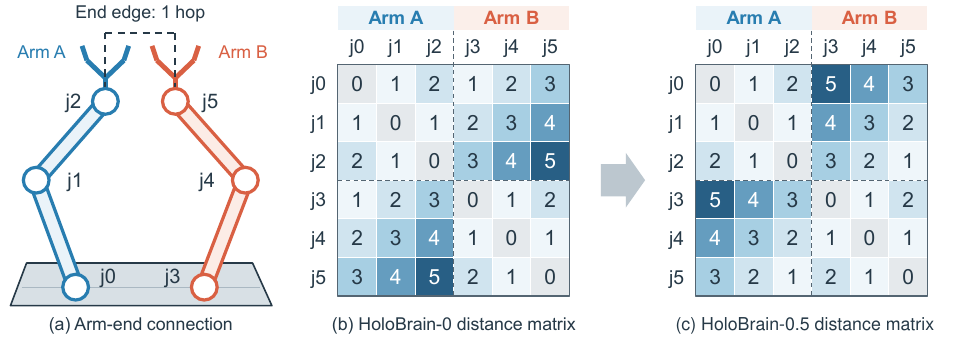}
    \caption{Joint-relative position encoding for an illustrative dual-arm embodiment. Entries give shortest-path hop counts on two three-joint chains with unit within-arm edges. (a)~The dashed connection joins $j_2$ and $j_5$ in HoloBrain-0.5. HoloBrain-0 (b)~joins $j_0$ and $j_3$ with one hop, whereas HoloBrain-0.5 (c)~replaces this edge with the one-hop $j_2$--$j_5$ connection. The distal-joint distance changes from five hops to one; within-arm distances are unchanged.}
    \label{fig:joint_relative_position}
\end{figure*}

\textbf{Modified joint-relative position encoding.} Joint self-attention relies on a joint-relative position matrix, whose entries measure the kinematic separation between joint pairs. In HoloBrain-0, these entries are hop counts on the kinematic graph: adjacent joints within an arm are one hop apart, and the two arms of a dual-arm robot are connected at their bases by a single unit-hop edge. The two end effectors are therefore separated by nearly the combined length of both chains, even though their actions may be strongly correlated. HoloBrain-0.5 instead connects the two arm ends with a single unit-hop edge, so every cross-arm distance becomes the shortest path from one joint to the end of its own arm, across this edge, and on to the other joint, while the within-arm kinematic topology is preserved. The original and modified matrices are compared in Figure~\ref{fig:joint_relative_position}.

\subsection{HoloBrain-0.5-Discrete}

HoloBrain-0.5-Discrete uses the same canonical joint-centric representation and multimodal conditioning interface as the continuous variant, but replaces direct action generation with categorical prediction. Training is divided into two stages. We first learn a morphology-aware action tokenizer that compresses continuous trajectories into discrete codes and reconstructs them without image or language input. We then freeze the tokenizer and train the VLA policy to predict its codes from visual observations, language instructions, and proprioceptive state. Because tokenizer parameters are shared and code indices come from a fixed alphabet, every embodiment's trajectories map into the same categorical output space even though the number of modeled joints or links varies.

\begin{figure}[!htbp]
    \centering
    \includegraphics[width=\linewidth]{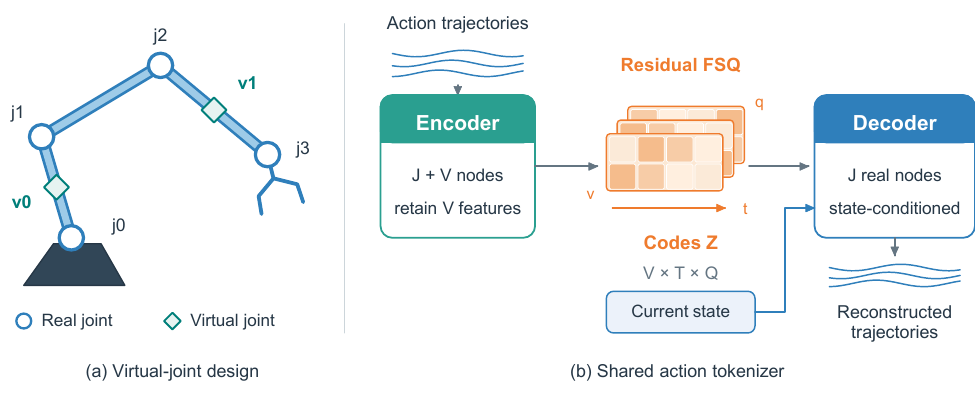}
    \caption{Virtual-joint design and action tokenization. (a)~Virtual nodes share a learned initialization and obtain graph-relative identities from their positions on the kinematic graph. (b)~The encoder processes real and virtual nodes jointly and retains the virtual-node features. Residual FSQ produces a code grid indexed by virtual joint $v$, temporal patch $t$, and residual stage $q$. State-conditioned decoding and restoration reconstruct action trajectories. Network parameters are shared across embodiment-dependent node counts. Trajectories and grid dimensions are schematic; implementation settings appear in Appendix~\ref{app:discrete-implementation}.}
    \label{fig:virtual-joints}
\end{figure}

\textbf{Virtual joints.} Let $\mathbf{A}\in\mathbb{R}^{H\times J\times d}$ be a future action trajectory, where $H$ is the prediction horizon and $J$ is the embodiment-dependent number of represented nodes. Each node contributes $d$ channels: a joint or gripper scalar and the pose of its link. The trajectory is divided into $T$ non-overlapping temporal patches. To summarize trajectories using their graph structure, we introduce latent aggregation nodes, termed \emph{virtual joints}, and place them on the kinematic graph (Figure~\ref{fig:virtual-joints}). For a motion chain $c$ with $n_c$ declared link and finger entries, we allocate
\begin{equation}
    V_c=\left\lfloor n_c/\rho\right\rfloor+1,\qquad V=\sum_c V_c,
    \label{eq:virtual-count}
\end{equation}
where $\rho$ controls compression; all tokenizer parameters are shared across these embodiment-dependent counts. Virtual joints are positioned to minimize the maximum distance from a real joint to its nearest virtual joint (Appendix~\ref{app:discrete-placement}). Each virtual joint's graph position determines its distances to the real nodes and to the other virtual joints. Adding these entries to the hop distances among the real nodes (Figure~\ref{fig:joint_relative_position}) yields the augmented matrix $\Delta\in\mathbb{R}^{(J+V)\times(J+V)}$, which conditions the tokenizer's attention.

\textbf{Graph aggregation and discretization.} Each temporal patch of a real-node trajectory is embedded into a feature. We append $V$ virtual nodes initialized from a shared learned vector and process the $J+V$ node streams with graph attention over the augmented distances $\Delta$. The attention itself is inherited from HoloBrain-0~\cite{holobrain0}: a sinusoidal embedding and MLP map each distance $\Delta_{ij}$ to a channel-wise gate that modulates query--key interactions, with temporal rotary encoding supplying time dependence. We retain the virtual-node features $\mathbf{H}\in\mathbb{R}^{V\times T\times C}$ as the compression bottleneck, where $C$ is the feature width.

We apply finite scalar quantization~\cite{fsq} to each bottleneck feature $\mathbf{H}_{vt}$ through $Q$ residual stages: the first stage quantizes the feature, and each subsequent stage quantizes the remaining residual at a successively finer scale. The indices emitted by the $Q$ stages form the code grid $\mathbf{Z}\in\{0,\ldots,K-1\}^{V\times T\times Q}$, indexed by virtual joint $v$, temporal patch $t$, and residual stage $q=0,\ldots,Q-1$, where $K$ is the number of indices per stage. Appendix~\ref{app:discrete-quantization} details the quantization configuration and implementation.

\textbf{Reconstruction.} Real-node queries initialized from the current proprioceptive state attend to the quantized virtual features. The decoder upsamples $T$ patches to $H$ frames and predicts changes in joint values and link poses relative to the current state. Applying these changes to that state recovers absolute joint and link-pose trajectories. Gripper openings are predicted directly as absolute values. Training every residual prefix encourages earlier stages to support coarse reconstruction and later stages to refine it. Let $\widehat{\mathbf{A}}^{(r)}$ be the reconstructed trajectory using the first $r$ stages. Tokenizer training minimizes
\begin{equation}
    \mathcal{L}_{\mathrm{tok}}
    = \sum_{r=1}^{Q}\lambda_r\,\mathcal{L}_{\mathrm{rec}}\!\left(\widehat{\mathbf{A}}^{(r)},\mathbf{A}\right),
    \label{eq:discrete_tokenizer_loss}
\end{equation}
where $\lambda_r$ weights the reconstruction loss of the $r$-stage prefix, and $\mathcal{L}_{\mathrm{rec}}$ compares joint scalars, positions, and rotations, with forward-kinematics supervision and pose-consistency terms where kinematics are available (Appendix~\ref{app:discrete-losses}). Figure~\ref{fig:virtual-joints}(b) summarizes the tokenizer's structure.

\begin{figure}[!htbp]
    \centering
    \includegraphics[width=\linewidth]{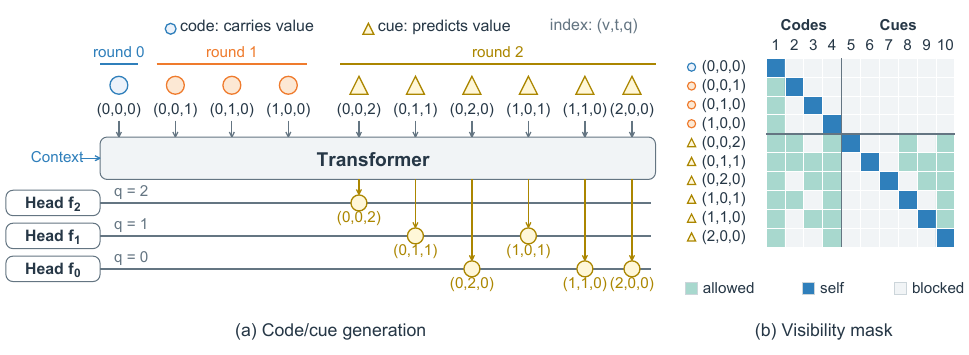}
    \caption{Code/cue delayed generation. (a)~Round-2 snapshot of a grid with $V,T,Q\geq 3$ and $\delta=t+v+q$: four known codes (circles, rounds 0--1) and six prediction cues (triangles, round 2). Each cue uses its residual-stage head $f_q$, shared across virtual joints and patches. (b)~Visibility among these positions: rows query columns; teal allows attention, blue marks self-attention, and gray blocks it. Positions 1--4 carry codes; 5--10 are cues whose target values remain unavailable. Figure~\ref{fig:delay-mask-components} decomposes the mask.}
    \label{fig:delayed_ar}
\end{figure}

\Needspace{6\baselineskip}
\textbf{Code/cue delayed autoregression.} The tokenizer represents each action chunk with codes for multiple virtual joints, temporal patches, and residual stages. Generating every code serially preserves all dependencies but requires $VTQ$ sequential policy evaluations; predicting them all in one pass removes this cost but cannot condition on code values sampled elsewhere in the same chunk. We instead predict groups of codes together, using permitted earlier predictions as context, following the delayed-codebook principle of~\cite{musicgen} (Figure~\ref{fig:delayed_ar}).

Each cell $i=(v,t,q)$ occupies two positions in the transformer sequence: a \emph{code position} carrying its value $Z_i$, and a \emph{cue position} from which that value is predicted. Their initial features are $\mathbf{h}_i^{\mathrm{code},0}=e(Z_i)+e_q^{\mathrm{code}}$ and $\mathbf{h}_i^{\mathrm{cue},0}=e_q^{\mathrm{cue}}$, where $e$ is a learned embedding lookup for code values and $e_q^{\mathrm{code}}$, $e_q^{\mathrm{cue}}$ are learned embeddings for residual stage $q$. Temporal rotary encoding identifies the patch, and graph distances identify the virtual joint's relationships to other nodes. The output feature at each cue gives $p_{\theta,i}=\operatorname{softmax}(f_q(\mathbf{h}_i^{\mathrm{cue}}))$, where $f_q$ is the classification head for residual stage $q$, shared across virtual joints and patches.

We advance diagonally through the code grid: cells whose virtual-joint, temporal-patch, and residual-stage indices have the same sum are predicted together, and each such diagonal defines a delay step. Using zero-based indices, we assign each cell a delay $\delta_i$ and collect the cells predicted at step $\ell$ in $\mathcal{G}_\ell$; a one-way mask $B$ additionally prevents conditioning on a later time or a finer residual stage even when its delay is smaller:
\begin{equation}
    \delta_i=t+v+q,\quad
    \mathcal{G}_\ell=\{i:\delta_i=\ell\},\quad
    B(i,j)=\mathbf1[t'\leq t]\,\mathbf1[q'\leq q],
    \label{eq:delay}
\end{equation}
where $j=(v',t',q')$ is a key cell. Let $\mathcal{P}(i)=\{j:\delta_j<\delta_i,\ B(i,j)=1\}$ denote the permitted earlier code positions. Given the multimodal context $\mathbf{c}=(o_h,i_h,s_h)$, comprising the observation, instruction, and current proprioceptive state at control step $h$, the policy factorizes as
\begin{equation}
    p_{\theta}(\mathbf{Z}\mid\mathbf{c})
    = \prod_{\ell}\prod_{i\in\mathcal{G}_{\ell}}
    p_{\theta}\!\left(Z_i\mid \mathbf{Z}_{\mathcal{P}(i)},\mathbf{c}\right).
    \label{eq:delayed_ar}
\end{equation}
Each code or cue position at $i$ reads $\mathcal{P}(i)$ and itself; cues also read same-step cues under $B$.

\begin{figure}[!htbp]
    \centering
    \includegraphics[width=\linewidth]{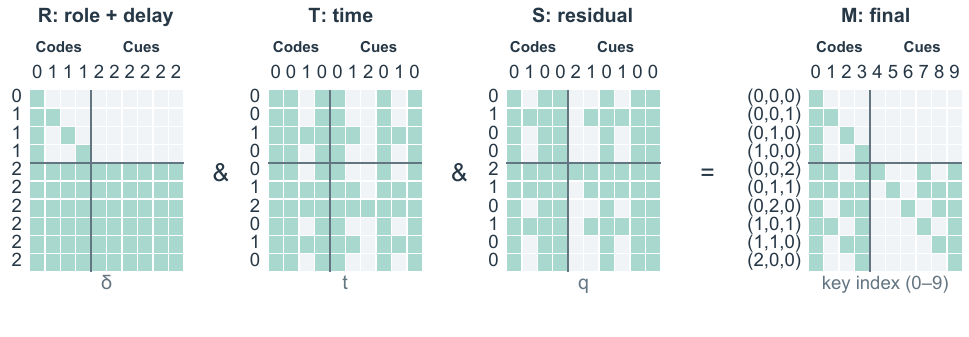}
    \caption{Factorization of the visibility mask in Figure~\ref{fig:delayed_ar}(b). Rows are queries and columns are keys, in the same position order. $R$ applies the code/cue role and delay rules, including self-attention; $T$ admits keys at time $t'\leq t$; $S$ admits keys at residual stage $q'\leq q$. Their intersection gives the final visibility $M$.}
    \label{fig:delay-mask-components}
\end{figure}

Figure~\ref{fig:delay-mask-components} decomposes the round-2 visibility of Figure~\ref{fig:delayed_ar}(b) into its role/delay, time, and residual components. The schedule therefore generates $VTQ$ codes in $T+V+Q-2$ rounds while retaining causal dependencies across temporal patches, virtual-joint summaries, and residual quantization stages.

With the tokenizer frozen, the action expert is trained by cross-entropy over the ground-truth grid (Appendix~\ref{app:discrete-losses}),
\begin{equation}
    \mathcal{L}_{\mathrm{AR}}
    = -\sum_{i}\log p_{\theta}\!\left(Z_i\mid\mathbf{Z}_{\mathcal{P}(i)},\mathbf{c}\right).
    \label{eq:discrete_ar_loss}
\end{equation}
At inference, cues predict each round's codes from codes sampled in earlier rounds. The completed code grid is mapped back to quantized features by residual FSQ, and the state-conditioned tokenizer decoder reconstructs a continuous action chunk.

\subsection{Data Recipe}
\begin{figure*}[!htbp]
    \centering
    \includegraphics[width=0.99\linewidth]{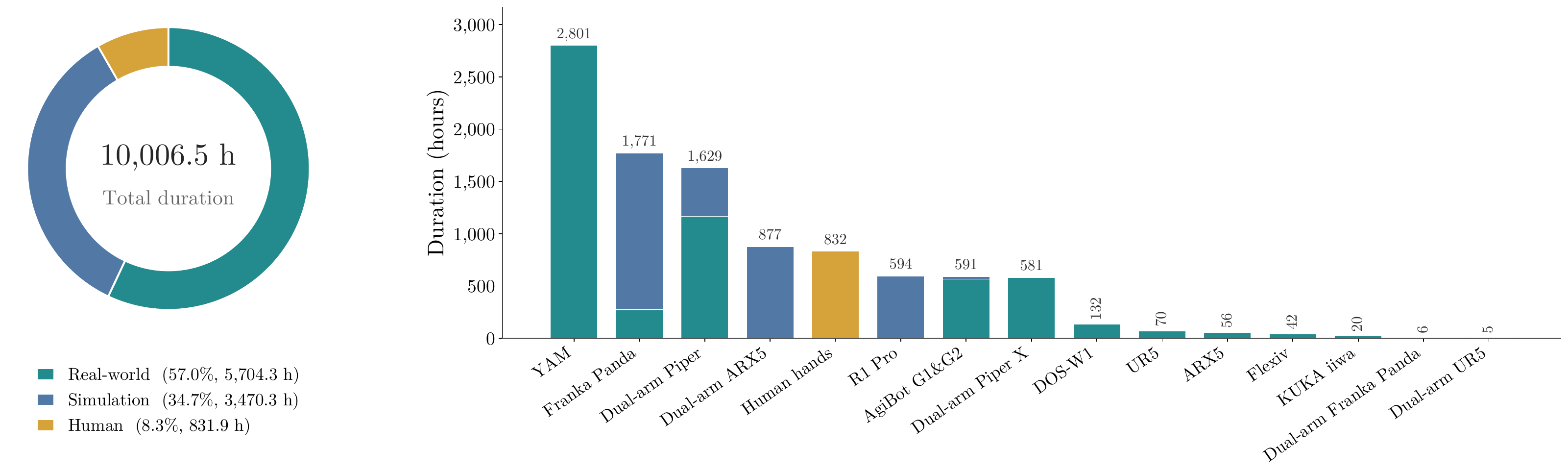}
    \caption{Composition of the HoloBrain-0.5 pretraining corpus. The left plot shows the duration contributed by real-robot, simulation, and human-manipulation data; the right plot breaks the same corpus down by embodiment category.}
    \label{fig:data_recipe}
\end{figure*}

Compared with HoloBrain-0, we scale the pretraining corpus from approximately 3,500 to 10,006.5 hours and increase its embodiment coverage from 7 to 15 categories. The corpus contains 5,704.3 hours of real-robot data, 3,470.3 hours of simulation, and 831.9 hours of human-manipulation data. Figure~\ref{fig:data_recipe} and Table~\ref{tab:data_recipe} summarize its composition.

A key property of the corpus is its consistent 3D geometric metadata. Robot trajectories include transformations between the robot and all cameras together with standardized URDF descriptions. For human-manipulation trajectories, we use tracked 3D hand joints to construct a gripper abstraction and define its pose directions according to the same convention used for robot grippers. This procedure places robot and human-hand data in a consistent 3D action representation. For open-source datasets that lack reliable camera calibration, we recover the missing parameters through manual correspondence annotation followed by bundle adjustment~\cite{bundleadjustment}.

Annotators mark physically identical landmarks across frames and camera views. Given manually initialized camera parameters, the corresponding 3D landmarks are initialized by multi-view triangulation. Camera intrinsics, camera poses, and landmark positions are then jointly refined by minimizing reprojection error:
\begin{equation}
    \min_{\{\boldsymbol{\theta}_c,\,\mathbf{T}_c,\,\mathbf{X}_k\}}
    \sum_{c,k}
    \left\|
    \mathbf{u}_{c,k}
    -
    \pi\left(
    \boldsymbol{\theta}_c,
    \mathbf{T}_c,
    \mathbf{X}_k
    \right)
    \right\|_2^2,
\end{equation}
where $\mathbf{u}_{c,k}$ is the annotated image location of landmark $k$ in camera $c$, $\mathbf{X}_k$ is its 3D position, $\boldsymbol{\theta}_c$ denotes camera intrinsics, $\mathbf{T}_c$ denotes camera extrinsics, and $\pi(\cdot)$ is the projection function. One calibrated camera or robot frame is fixed to remove gauge ambiguity. The optimized camera system is then aligned with the robot coordinate frame using known geometric constraints, and all results are manually checked by multi-view reprojection before entering the training corpus.

\begin{table*}[!htbp]
  \centering
  \caption{HoloBrain-0.5 pretraining data by dataset family. FPS is the effective frame rate, rounded to the nearest whole number; family and total durations are rounded independently to the nearest hour.}
  \label{tab:data_recipe}
  \small
  \setlength{\tabcolsep}{2pt}
  \renewcommand{\arraystretch}{1.05}
  \begin{tabular}{@{}>{\raggedright\arraybackslash}p{1.9cm}>{\raggedright\arraybackslash}p{3.35cm}>{\raggedright\arraybackslash}p{6.6cm}rr@{}}
    \toprule
    Data type & Dataset name & Embodiment & FPS & Duration (h) \\
    \midrule
    \multirow[t]{6}{*}{Real-world} & \texttt{Horizon dataset} & Dual-arm Piper / Dual-arm Piper X & 30 & 1,653 \\
     & \texttt{ABC-130K}~\cite{abc130k} & YAM & 30 & 2,801 \\
     & \texttt{AgiBot}~\cite{agibotworld} & AgiBot G1\&G2 & 30 & 564 \\
     & \texttt{DROID}~\cite{droid} & Franka Panda & 14 & 265 \\
     & \texttt{RH20T}~\cite{rh20t} & Flexiv / Franka Panda / KUKA iiwa / UR5 & 10 & 89 \\
     & \texttt{Table30}~\cite{robochallenge} & Dual-arm Piper / ARX5 / DOS-W1 / UR5 & 28 & 331 \\
    \midrule
    \addlinespace[1pt]
    \multirow[t]{8}{*}{Simulation} & \texttt{Genie Sim 3.0}~\cite{geniesim3} & AgiBot G1\&G2 & 30 & 26 \\
     & \texttt{BEHAVIOR-1K}~\cite{behavior1k} & R1 Pro & 30 & 594 \\
     & \texttt{Isaac Pick-Place} & Franka Panda / Dual-arm Piper X & 30 & 1 \\
     & \texttt{LIBERO}~\cite{libero} & Franka Panda & 20 & 4 \\
     & \texttt{RoboCasa}~\cite{robocasa} & Franka Panda & 20 & 1,495 \\
     & \texttt{RoboDojo}~\cite{robodojo} & Dual-arm ARX5 & 25 & 21 \\
     & \texttt{RoboTwin}~\cite{robotwin,robotwin2} & Dual-arm ARX5 / Dual-arm Franka Panda / Dual-arm Piper / Dual-arm UR5 & 17 & 170 \\
     & \texttt{InternData-A1}~\cite{interndata} & Dual-arm Piper / Dual-arm ARX5 & 30 & 1,159 \\
    \midrule
    \addlinespace[1pt]
    \multirow[t]{1}{*}{Human} & \texttt{EgoDex}~\cite{egodex} & Human hands & 30 & 832 \\
    \midrule
    \multicolumn{3}{r}{\textbf{Total}} & -- & \textbf{10,007} \\
    \bottomrule
  \end{tabular}
\end{table*}

\section{Experiments}

We track closed-loop performance throughout pretraining and evaluate the resulting policies before and after task-specific post-training. Our real-robot evaluation spans five embodiments: Piper, Piper-X, a single-arm Franka Panda, a mobile dual-arm platform, and a Unitree G1 humanoid. We distinguish two forms of cross-embodiment evaluation: transfer without demonstrations of the target task on the target robot, and adaptation using a limited set of demonstrations from the target embodiment.

\subsection{Closed-Loop Performance of Pretrained Models}

\begin{figure*}[!htbp]
    \centering
    \includegraphics[width=0.99\linewidth]{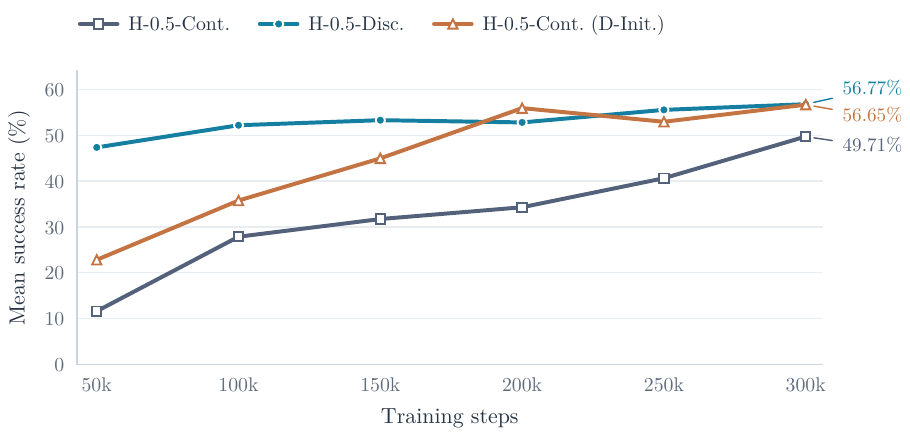}
    \caption{Closed-loop simulation performance throughout pretraining. Each model checkpoint is evaluated over 21,430 episodes in total across six simulation settings spanning three simulators. The plotted metric is the unweighted mean of the six setting-specific success rates.}
    \label{fig:pretraining_curve}
\end{figure*}

We evaluate each pretrained checkpoint over 21,430 episodes across six simulation settings: three 50-task RoboTwin settings (clean, randomized, and UR5-WSG)~\cite{robotwin2}, LIBERO~\cite{libero}, LIBERO-Plus~\cite{liberoplus}, and RoboCasa~\cite{robocasa}. Figure~\ref{fig:pretraining_curve} reports the unweighted mean of the six per-setting success rates.

{\emergencystretch=2em
After pretraining, HoloBrain-0.5-Continuous achieves a mean simulation success rate of 49.71\%, while HoloBrain-0.5-Discrete reaches 56.77\%. The discrete variant reaches 47.34\% at 50,000 steps and remains above the continuous curve at the plotted checkpoints.\par
}

We further train an additional continuous variant by initializing its shared components with the parameters of HoloBrain-0.5-Discrete and replacing its discrete action head with a diffusion head. We denote this model HoloBrain-0.5-Continuous (D-Init.). This variant achieves higher success rates than the continuous baseline at the plotted checkpoints, suggesting that discrete-action pretraining can provide a useful initialization for continuous action generation.

In addition to simulation-based evaluation, we conduct out-of-the-box real-robot experiments with all three pretrained models. The corresponding results are summarized in Table~\ref{tab:real_robot}.

\begin{table}[!htbp]
    \centering
    \caption{Out-of-the-box and post-trained real-robot performance, measured by progress score (\%). The upper rows report pretrained models evaluated without task-specific post-training; rows marked with $\checkmark$ report task-specific post-trained models. Bold scores mark the best results within each training setting, including ties. H-0.5-Cont. and H-0.5-Disc. denote HoloBrain-0.5-Continuous and HoloBrain-0.5-Discrete, respectively, and D-Init. denotes initialization from HoloBrain-0.5-Discrete.}
    \label{tab:real_robot}
    \setlength{\tabcolsep}{4pt}

    \resizebox{\linewidth}{!}{
    \begin{tabular}{llcccc}
        \Xhline{1.5pt}
        Post-train &
        Model
        & Place shoes
        & Fold towel
        & Stack three blocks
        & Place bottles in dustbin \\
        \midrule

        & H-0.5-Cont.
        & 91.00 & 63.20 & 64.44 & \textbf{91.67} \\

        & H-0.5-Disc.
        & \textbf{96.50} & \textbf{81.60} & \textbf{82.22} & 84.76 \\

        & H-0.5-Cont. (D-Init.)
        & 91.60 & 76.30 & 72.22 & 86.90 \\

        \midrule
        \checkmark & $\pi_{0}$ & 75.48 & 61.58 & 70.00 & 55.24 \\
        \checkmark & $\pi_{0.5}$ & 84.19 & 61.58 & 80.00 & 94.29  \\
        \checkmark & HoloBrain-0 & 93.23 & 84.74 & 83.33 & 95.24 \\
        \checkmark & H-0.5-Cont. &  \textbf{98.40}	& \textbf{97.40}	& \textbf{100.00}	& \textbf{100.00}	\\
        \checkmark & H-0.5-Disc. &  95.50& 	71.10	& 57.78	& 97.62	\\
        \checkmark & H-0.5-Cont. (D-Init.) & \textbf{98.40} & 	89.50 & \textbf{100.00}	& \textbf{100.00}\\

        \addlinespace
        \midrule
        \addlinespace

        Post-train &
        Model
        & Place empty cup
        & Place in slot
        & Stack three bowls
        & Average \\
        \midrule

        & H-0.5-Cont.
        & 73.50 & 77.33 & 96.44 & 79.65 \\

        & H-0.5-Disc.
        & \textbf{91.00} & 80.89 & \textbf{97.33} & \textbf{87.76} \\

        & H-0.5-Cont. (D-Init.)
        & 77.00 & \textbf{85.78} & 95.56 & 83.62 \\
        \midrule
        \checkmark & $\pi_0$
& 48.50 & 68.00 & 97.78 & 68.08 \\

\checkmark & $\pi_{0.5}$
& \textbf{99.50} & 84.22 & \textbf{100.00} & 86.25 \\

\checkmark & HoloBrain-0
& 78.50 & 62.22 & \textbf{100.00} & 85.32 \\

\checkmark & H-0.5-Cont.
& 90.50 & 78.44 & 96.89 & \textbf{94.52} \\

\checkmark & H-0.5-Disc.
& 37.50 & 30.44 & 98.22 & 69.74 \\

\checkmark & H-0.5-Cont. (D-Init.) & 60.50& 	\textbf{85.78}	& 99.56 & 90.53 \\
        \bottomrule
    \end{tabular}
    }
\end{table}

\subsection{Post-Training Performance}

We evaluate the post-training performance of HoloBrain-0.5 under two settings that differ in data scale and task diversity.

In the first setting, we post-train the models on seven real-robot tasks using a single embodiment, with 200 demonstration trajectories provided for each task. This controlled setting evaluates adaptation under a limited data budget. As shown in Table~\ref{tab:real_robot}, HoloBrain-0.5-Continuous achieves an average progress score of 94.52\% after post-training, outperforming $\pi_{0.5}$ and HoloBrain-0 by 8.27 and 9.20 percentage points, respectively.

Under the same few-shot fine-tuning setting, the average progress score of HoloBrain-0.5-Discrete decreases from 87.76\% to 69.74\%. The current discrete model is therefore not well suited to few-shot post-training. We will investigate this performance degradation in future work.

In the second setting and in the transfer and adaptation experiments below, we use the continuous policy initialized from the discrete model (D-Init.). We perform large-scale multi-task post-training using approximately 300 hours of data collected with two dual-arm robot embodiments, Piper and Piper-X. The complete dataset covers 38 tasks, from which we select 11 representative tasks for evaluation. Depending on task complexity, the amount of training data assigned to each task ranges from 2 to 40 hours, with more challenging tasks generally receiving larger data budgets. This setting evaluates the ability of HoloBrain-0.5 to leverage large-scale heterogeneous data and adapt to more complex real-world manipulation tasks. The results are presented in Table~\ref{tab:large_scale_post_training}.

\begin{table*}[!htbp]
    \centering
    \caption{
        Large-scale multi-task post-training performance on real-robot
        tasks, measured by progress score (\%). H-0.5-Cont. (D-Init.)
        denotes HoloBrain-0.5-Continuous initialized from HoloBrain-0.5-Discrete. In Tables~\ref{tab:large_scale_post_training}--\ref{tab:adaptation-g1}, the green number is the lead over the second-best average, in percentage points.
    }
    \label{tab:large_scale_post_training}
    \setlength{\tabcolsep}{5pt}
    \renewcommand{\arraystretch}{1.15}

    \resizebox{\textwidth}{!}{%
    \begin{tabular}{l|cccccc}
        \toprule
        Model
        & Grasp anything
        & Fold paper box
        & Object to cabinet
        & Object to drawer
        & Objects to bag
        & Sweep trash \\
        \midrule

         $\pi_0$
        & 88.00 & 65.50 & 73.00 & 70.50 & 84.40 & 57.50 \\
        $\pi_{0.5}$
        & \textbf{99.00} & 68.00 & 74.80 & 68.30 & \textbf{91.60} & 42.50 \\
        H-0.5-Cont. (D-Init.)
        & 97.50 & \textbf{81.00} & \textbf{84.50} & \textbf{96.00} & 86.70 & \textbf{72.00} \\
        \midrule
        Model
        & Close laptop
        & Pen to holder
        & Hard slot
        & Phone to stand
        & Drawstring bag
        & Average \\
        \midrule
        $\pi_0$
        & 46.70 & 16.70 & 56.67 & \textbf{73.10} & 56.50 & 62.60 \\
        $\pi_{0.5}$
        & 46.70 & 66.70 & 78.00 & 64.40 & \textbf{63.50} & 69.41 \\
        H-0.5-Cont. (D-Init.)
        & \textbf{66.70} & \textbf{88.00} & \textbf{82.00} & 66.90 & 48.50 & \textbf{79.07}\scoregain{9.66} \\
        \bottomrule
    \end{tabular}%
    }
\end{table*}

\subsection{Cross-Embodiment Generalization}
\label{sec:zero-shot-transfer}

We evaluate zero-shot task transfer from Piper to Piper-X on five real-robot tasks. The model has encountered Piper-X on other tasks, but the five evaluated task--Piper-X combinations are absent from both pretraining and post-training data. Training demonstrations for these five tasks come only from Piper. This setting tests generalization to unseen task--embodiment combinations on a previously seen embodiment.

Piper and Piper-X use identical sensor configurations and have the same image and joint-state dimensionalities, so input dimensions alone do not identify the embodiment. Models without an explicit embodiment condition, such as the standard $\pi$-series configurations used here, do not receive a direct signal specifying the target robot's kinematic structure. HoloBrain-0.5 conditions action generation on standardized URDF and joint-level kinematic information, providing an explicit route for adapting the learned skill to the target embodiment.

The results are summarized in Table~\ref{tab:cross_embodiment}. We report the progress score for each task and the average across the five tasks. This evaluation measures whether skills learned from Piper demonstrations transfer to Piper-X without demonstrations of the evaluated task--Piper-X combinations, despite differences in kinematic structure.

\begin{table*}[!htbp]
    \centering
    \caption{Zero-shot task transfer from Piper to Piper-X, measured by progress score (\%). Piper-X is seen on other tasks, but the evaluated task--Piper-X combinations are absent from both pretraining and post-training data. All models are post-trained using Piper demonstrations only and evaluated directly on Piper-X.}
    \label{tab:cross_embodiment}
    \setlength{\tabcolsep}{4pt}
    \resizebox{\textwidth}{!}{%
    \begin{tabular}{l|ccccc|c}
        \toprule
        Model
        & Object in bag
        & Object to drawer
        & Close laptop
        & Objects to basket
        & Phone to stand
        & Average \\
        \midrule
        $\pi_0$ & 9.40 & 72.00 & 26.70 & 2.00 & 28.10 & 27.64 \\
        $\pi_{0.5}$ & 1.90 & 47.00 & 13.30 & 2.50 & 22.50 & 17.44 \\
        H-0.5-Cont. (D-Init.) & \textbf{38.10} & \textbf{91.00} & \textbf{53.30} & \textbf{91.50} & \textbf{42.50} & \textbf{63.28}\scoregain{35.64} \\
        \bottomrule
    \end{tabular}%
    }
\end{table*}

\subsection{Adaptation to Different Robot Morphologies}
\label{sec:embodiment-adaptation}

The Piper-to-Piper-X experiment in Section~\ref{sec:zero-shot-transfer} asks whether a policy can transfer a task without seeing it demonstrated on the target robot. Here we ask how well the policy adapts when a small amount of target-embodiment data is available. We consider three distinct configurations: a single-arm Franka Panda, a mobile base with two manipulator arms, and a Unitree G1 humanoid. For each configuration, the comparison uses the same set of target-embodiment demonstrations to post-train HoloBrain-0.5-Continuous and $\pi_{0.5}$. We compare the resulting policies within each embodiment using task progress scores.

For the single-arm Franka, we post-train both models exclusively on the same subset of DROID~\cite{droid} demonstrations. Neither model is fine-tuned on data collected in our evaluation environment or on demonstrations collected specifically for the five tabletop evaluation tasks. On these tasks (Table~\ref{tab:embodiment-adaptation}), HoloBrain-0.5-Continuous attains a mean progress score of 89.2\%, compared with 85.0\% for $\pi_{0.5}$. It scores higher on four of the five tasks. The largest difference is on placing a marker in a cup (85.0\% versus 77.5\%); on cup placement, $\pi_{0.5}$ scores slightly higher (95.8\% versus 95.0\%).

The mobile dual-arm tasks (Table~\ref{tab:adaptation-mobile}) are pressing an elevator button, placing a basket, and hanging a garment on a rack. HoloBrain-0.5-Continuous achieves an average progress score of 92.94\% on these three tasks, compared with 73.83\% for $\pi_{0.5}$. The Unitree G1 tasks (Table~\ref{tab:adaptation-g1}) include retrieving a drink from a refrigerator and placing it on a table, placing a bag of dirty clothes on a table and returning, and handing a bag of clothes to a person. For the Unitree G1 experiments, we use HoloMotion-1~\cite{holomotion} as the whole-body motion tracker. HoloBrain-0.5-Continuous achieves an average progress score of 70.24\% on the three G1 tasks, compared with 11.00\% for $\pi_{0.5}$.

\begin{table*}[!htbp]
    \centering
    \caption{Single-arm Franka Panda progress scores (\%) after post-training on the same DROID subset. Neither model uses fine-tuning demonstrations collected in our evaluation environment or specifically for the five tasks. The average weights all five tasks equally.}
    \label{tab:embodiment-adaptation}
    \small
    \setlength{\tabcolsep}{3pt}
    \begin{tabular}{l|ccccc|c}
        \toprule
        Model & \makecell{Place a\\cup} & \makecell{Transfer a\\towel}
        & \makecell{Move a bowl\\to the left} & \makecell{Put an object\\on a plate}
        & \makecell{Put a marker\\in a cup} & Average \\
        \midrule
        $\pi_{0.5}$ & \textbf{95.8} & 77.5 & 92.5 & 81.7 & 77.5 & 85.0 \\
        H-0.5-Cont. (D-Init.) & 95.0 & \textbf{82.5} & \textbf{97.5} & \textbf{85.8} & \textbf{85.0} & \textbf{89.2}\scoregain{4.20} \\
        \bottomrule
    \end{tabular}
\end{table*}

\begin{table*}[!htbp]
    \centering
    \caption{Task progress scores (\%) on the mobile dual-arm robot after post-training on matched target-embodiment data. The average weights all three tasks equally.}
    \label{tab:adaptation-mobile}
    \small
    \setlength{\tabcolsep}{4pt}
    \begin{tabular}{l|ccc|c}
        \toprule
        Model & \makecell{Press an\\elevator button} & Place a basket
        & \makecell{Hang a garment\\on a rack} & Average \\
        \midrule
        $\pi_{0.5}$ & 80.00 & 62.50 & 79.00 & 73.83 \\
        H-0.5-Cont. (D-Init.) & \textbf{97.62} & \textbf{89.69} & \textbf{91.50} & \textbf{92.94}\scoregain{19.11} \\
        \bottomrule
    \end{tabular}
\end{table*}

\begin{table*}[!htbp]
    \centering
    \caption{Task progress scores (\%) on the Unitree G1 humanoid after post-training on matched target-embodiment data. The average weights all three tasks equally.}
    \label{tab:adaptation-g1}
    \small
    \setlength{\tabcolsep}{4pt}
    \begin{tabular}{l|ccc|c}
        \toprule
        Model & \makecell{Retrieve drink\\from fridge} & \makecell{Bag to table\\+ return}
        & \makecell{Bag to\\person} & Average \\
        \midrule
        $\pi_{0.5}$ & 23.00 & 10.00 & 0.00 & 11.00 \\
        H-0.5-Cont. (D-Init.) & \textbf{83.00} & \textbf{100.00} & \textbf{27.73} & \textbf{70.24}\scoregain{59.24} \\
        \bottomrule
    \end{tabular}
\end{table*}

\section{Conclusion}
\label{sec:conclusion}

We introduce HoloBrain-0.5, a family of VLA models that extends HoloBrain-0 in model architecture, pretraining scale, and evaluation methodology. First, we further unify heterogeneous action spaces through URDF standardization to support learning across embodiments with different joint configurations and kinematic structures. Second, we develop a discrete-action variant that exhibits faster convergence during pretraining and achieves stronger out-of-the-box performance. Finally, we pretrain HoloBrain-0.5 on more than 10,000 hours of heterogeneous manipulation data and systematically evaluate its performance throughout pretraining.
HoloBrain-0.5 represents a step toward scalable, embodiment-aware generalist manipulation policies. The results support the use of explicit kinematic canonicalization and complementary action-space formulations for learning from diverse robotic data.

\section{Future Work}
\label{sec:future_work}

HoloBrain-0.5 is currently an action-supervised VLA model designed primarily for reactive visuomotor control. We plan to extend its capabilities along four directions:

\begin{itemize}
    \item \textbf{Broader multi-task supervision.}
        We will incorporate a more diverse collection of tasks into joint training, including visual question answering with textual responses and 3D grounding. Such supervision may improve semantic understanding, spatial reasoning, and the connection between high-level instructions and physical actions.

    \item \textbf{Scaling up training data.}
        We will further expand the scale of training data. HoloBrain's current use of egocentric human data is exploratory; we plan to substantially increase its proportion in the training mixture and build a more balanced data pyramid.

    \item \textbf{Temporal modeling and memory.}
        HoloBrain-0.5 currently makes decisions from visual observations at the current time step. Future versions will incorporate memory mechanisms and temporal pretraining, enabling the model to retain and use historical observations.

    \item \textbf{Richer physical modalities.}
    Vision, language, and action alone provide an incomplete representation of physical interaction. We plan to incorporate additional sensory signals, including tactile observations and joint-level force/torque feedback.
\end{itemize}

\clearpage
\appendix
\section{HoloBrain-0.5-Discrete Details}
\label{app:discrete-details}

This appendix supplements Section~3.3 with the virtual-joint placement procedure, the residual-quantization configuration, loss and regularization details, and implementation settings.

\subsection{Virtual-Joint Placement}
\label{app:discrete-placement}

\begin{figure}[!htbp]
    \centering
    \includegraphics[width=0.85\linewidth]{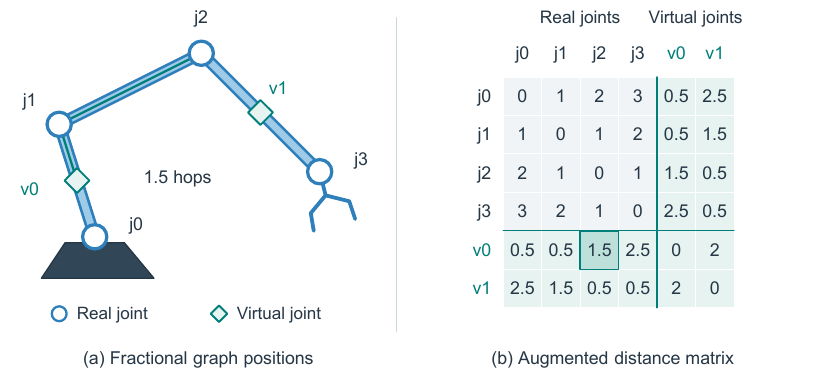}
    \caption{Illustrative distance augmentation for one motion chain. (a)~Four real nodes and two midpoint virtual nodes, with a highlighted $1.5$-hop path from $v_0$ to $j_2$. Each real-node edge has unit hop length. (b)~The augmented distance matrix contains real--real, real--virtual, and virtual--virtual entries; the highlighted $(v_0,j_2)$ entry matches the $1.5$-hop path in (a). This single-chain example illustrates fractional offsets. Cross-arm relationships in dual-arm embodiments follow Figure~\ref{fig:joint_relative_position}.}
    \label{fig:virtual-distance-details}
\end{figure}

The placement algorithm initializes $V$ virtual-joint positions on the kinematic graph and refines them locally. Both stages use a lexicographic objective: first minimize the maximum distance from a real joint to its nearest virtual joint, then the mean of these distances, and finally maximize the minimum pairwise distance between virtual joints. This objective favors coverage of the real joints while spreading the virtual joints along the chains. Fractional positions along physical edges extend the joint-relative relationships to real--virtual and virtual--virtual pairs, measured in physical-hop units; Figure~\ref{fig:virtual-distance-details} illustrates the augmented distances on a single motion chain. The real-to-real block uses the modified cross-arm relationships of Figure~\ref{fig:joint_relative_position} while retaining the within-arm distances.

\subsection{Residual Quantization Configuration}
\label{app:discrete-quantization}

The quantizer follows the open-source residual FSQ implementation in \texttt{vector-quantize-pytorch}\footnote{\url{https://github.com/lucidrains/vector-quantize-pytorch}}, which chains finite scalar quantization~\cite{fsq} stages residually. For each latent $\mathbf{h}=\mathbf{H}_{vt}$, a shared affine projection produces four quantization coordinates $u\in\mathbb{R}^{4}$, soft-bounded as $r_0=b\odot\tanh(u\oslash b)$ with $b=1+(L-1)^{-1}$ and $L=(8,5,5,5)$, all elementwise. Stage $q$ divides the residual by $s_q=L^{-q}$, quantizes each coordinate by scalar FSQ with straight-through rounding, rescales the quantized contribution $\widetilde c_q=s_q\odot c_q$, and passes on the residual $r_{q+1}=r_q-\operatorname{sg}(\widetilde c_q)$, with the subtracted contribution detached. The four level indices of a stage are packed with the mixed-radix basis $(1,8,40,200)$ into a single index $z_{vtq}\in\{0,\ldots,999\}$, so $K=1000$. An $r$-stage prefix is decoded by summing its contributions and applying the shared output projection once, $\overline{\mathbf{h}}^{(r)}=W_{\mathrm{out}}\sum_{q<r}\widetilde c_q+b_{\mathrm{out}}$.

\subsection{Losses and Regularization}
\label{app:discrete-losses}

Let $\ell_w(\mathbf{B},\mathbf{A})$ sum weighted smooth-$L_1$ errors on node scalars, positions, and rotation-matrix entries, using $\beta=10^{-4}$ and dataset-dependent channel weights. Errors are summed over node and channel entries and averaged over valid frames; a temporal patch is valid only if all its frames are valid. The reconstructed absolute trajectory and the target are compared in the same space: normalized scalars and link positions in the robot's metric frame, with quaternion outputs converted to $3\times3$ rotation matrices for the rotation term. When kinematics are supplied, define $F(\mathbf{B})$ to preserve $\mathbf{B}$'s scalar channels and replace its pose channels with forward kinematics of the predicted joint scalars, expressed in the same loss space. The reconstruction objective in Eq.~\eqref{eq:discrete_tokenizer_loss} is
\begin{equation}
\mathcal L_{\mathrm{rec}}(\mathbf{B},\mathbf{A})=\ell_w(\mathbf{B},\mathbf{A})+
\mathbf{1}_{\mathrm{FK}}\left[\ell_w(F(\mathbf{B}),\mathbf{A})+\ell_w\bigl(\mathbf{B},\operatorname{sg}(F(\mathbf{B}))\bigr)\right],
\label{eq:full-rec}
\end{equation}
where the forward-kinematics term differentiates through the kinematic model to the predicted scalars, and the consistency term pulls predicted poses toward detached forward-kinematics poses. When kinematics are absent, only the direct term is used.

The policy is trained on ground-truth codes with corrupted inputs, label smoothing, and residual-stage masks:
\begin{equation}
\mathcal L_{\mathrm{policy}}=
\mathbb E_{\widetilde Z\mid Z}\!\left[
\sum_{q=0}^{Q-1}\alpha_q
\frac{\sum_{v,t} M_{v,t,q}\,\operatorname{CE}_{\epsilon}\!\left(z_{v,t,q},\,p_{\theta,v,t,q}(\widetilde Z;\mathbf{c})\right)}
{\max\!\left(1,\sum_{v,t}M_{v,t,q}\right)}
\right]+\mathcal L_{\mathrm{depth}},
\label{eq:policy-loss}
\end{equation}
where $M$ combines valid-patch and cascading corruption masks, $\epsilon=0.1$, and $\alpha=(1,0.8,0.6)$. Input codes are independently replaced by uniform indices with stage probabilities $(0.2,0.5,0.7)$ while targets remain unchanged; for a given $(v,t)$, corruption at stage $q$ removes loss terms at later residual stages through
\begin{equation}
M_{v,t,q}=m_{v,t}\prod_{k=0}^{q-1}(1-U_{v,t,k}),
\end{equation}
where $m_{v,t}$ marks a valid temporal patch and $U_{v,t,k}$ marks an input-code replacement. The policy additionally drops state conditioning with probability $0.05$ during training and inherits auxiliary visual depth supervision from HoloBrain-0.

\subsection{Implementation Settings}
\label{app:discrete-implementation}

Figure~\ref{fig:virtual-joints}(b) in Section~3.3 summarizes the tokenization and reconstruction pipeline, whose encoder and decoder use pre-normalization, residual connections, and a final RMSNorm~\cite{rmsnorm}.
Table~\ref{tab:discrete-settings} lists the default settings.

\begin{table}[!htbp]
\caption{Default tokenizer settings. One action chunk contains the full future trajectory; a temporal patch is the shorter input unit.}
\label{tab:discrete-settings}
\centering
\small
\begin{tabular}{ll}
\toprule
Setting & Value \\
\midrule
Action chunk length $H$ / current-state length & 64 / 1 frames \\
Frames per temporal patch / patch count $T$ & 4 / 16 \\
Hidden width $C$ / attention heads & 384 / 8 \\
Tokenizer encoder / decoder blocks & 4 / 4 \\
Virtual-joint allocation ratio $\rho$ & 6 \\
Residual stages $Q$ / indices per stage $K$ & 3 / 1000 \\
FSQ scalar levels & $(8,5,5,5)$ \\
Residual-prefix loss weights $\lambda$ & $(0.25,0.5,1)$ \\
\bottomrule
\end{tabular}
\end{table}

Distance encoding uses 256 sinusoidal features with maximum period 10{,}000 and input scale one; temporal rotary encoding uses base 10{,}000. The policy uses the same Qwen3.5 backbone as the continuous variant. Its robot-state encoder is initialized from the tokenizer and then trained independently while the tokenizer remains frozen. Each of the $Q$ classification heads is shared across virtual joints and temporal patches. For $V=4$, the default settings produce 192 code targets and 384 code/cue positions, generated in 21 rounds.

\section{Real-Robot Evaluation Protocol}
\label{app:evaluation_protocol}

All real-robot results are reported as progress scores normalized to $[0,100]\%$. Each task is decomposed into observable milestones, and the score is the normalized sum of the milestone weights. The protocol for the reported real-robot tasks is summarized in Table~\ref{tab:scoring_protocol}. Time limits and retry limits are included where they affect scoring.

\begin{table*}[!htbp]
    \centering
    \caption{Compact real-robot progress-scoring protocol. Scores are normalized to 100\%; the listed milestones are the primary components of each task score.}
    \label{tab:scoring_protocol}
    \small
    \setlength{\tabcolsep}{4pt}
    \renewcommand{\arraystretch}{1.08}
    \begin{tabularx}{\textwidth}{@{}>{\raggedright\arraybackslash}p{3.0cm}X>{\raggedright\arraybackslash}p{2.3cm}@{}}
        \toprule
        Task & Scoring milestones & Constraints \\
        \midrule
        Place shoes & Grasp shoe (.4); partial placement (.2); complete placement (.2); correct orientation (.2). & -- \\
        Fold towel & Corner grasp (.1); first fold (.3); opposite-side grasp (.1); second fold (.2); final quality (.2). & -- \\
        Stack three blocks & Per block: grasp (.25) and successful placement (.25). & -- \\
        Place bottles in dustbin & Per bottle: grasp (.3) and successful placement (.2). & -- \\
        Place empty cup & Grasp (.5); partial placement (.2); complete placement (.3). & -- \\
        Place in slot & Per block: grasp (.25); within 2 cm (.15); complete placement (.1). & At most 3 attempts per block. \\
        Stack three bowls & Per bowl: grasp (.25); partial stack (.15); complete stack (.1). & At most 3 attempts per bowl. \\
        \midrule
        Grasp anything & Per object: grasp and place (.1), or grasp only (.05). & At most 3 attempts per object. \\
        Fold paper box & Left grasp (.2); right grasp (.1); first fold (.3); second fold (.3); placement (.1). & -- \\
        Object to cabinet & Unlock (.2); open (.15); each object (.1); close (.15); relock (.2). & 3 min. \\
        Object to drawer & Reach/grasp handle (.15/.2); open (.1); each object (.1); close (.25). & Time limit. \\
        Objects to bag & Unzip (.3); open bag (.05); each object (.15); close zipper (.35, or .2 if partial). & -- \\
        Sweep trash & Grasp broom (.3) and dustpan (.2); sweep (.2); empty (.2); return tools (.1). & 3 min. \\
        Close laptop & Fully close the laptop lid (1.0). & Binary success. \\
        Pen to holder & Start from 1.0; subtract .1 for each failed attempt. & 3 min. \\
        Hard slot & Per item: grasp (.1); place (.4), with partial credit at 80\% overlap. & 25 s per item. \\
        Phone to stand & Position stand (.2); grasp phone (.4); place phone (.4, or .2 if imperfect). & -- \\
        Drawstring bag & Open (.3); each object (.1); grasp both cords (.2); pull closed (.3). & -- \\
        \midrule
        Retrieve drink from fridge & Open refrigerator (.3); grasp drink (.3); close refrigerator (.2); place drink (.1); stable placement (.1). & Each subtask limited to 30 s. \\
        Bag to table + return & Grasp clothes bag (.2); place on table (.6); return to starting position (.2). & Each subtask limited to 30 s. \\
        Bag to person & Walk to table (.1); grasp clothes bag (.7); return and hand bag to person (.2). & Each subtask limited to 30 s. \\
        \bottomrule
    \end{tabularx}
\end{table*}

\clearpage
\section{Joint-Graph Examples}
\label{app:joint-graph-examples}

\begin{figure}[!htbp]
    \centering
    \begin{subfigure}[t]{0.49\linewidth}
        \centering\includegraphics[width=\linewidth]{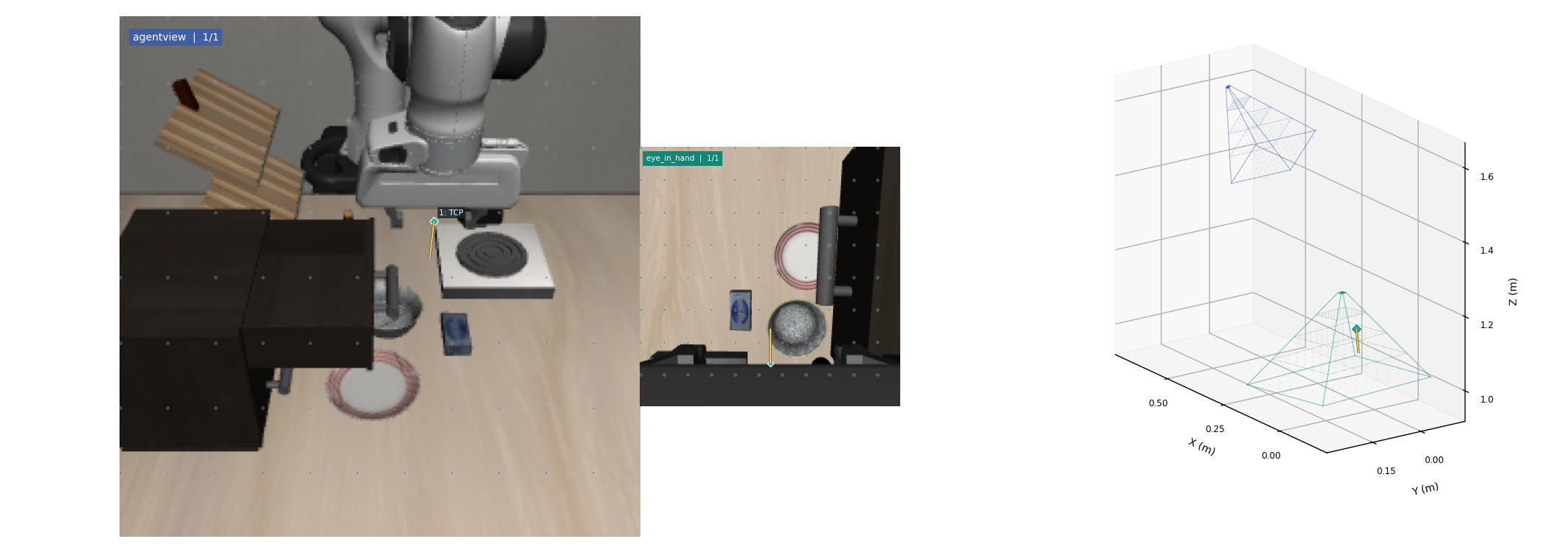}
        \caption{LIBERO Goal (TCP only)}
    \end{subfigure}\hfill
    \begin{subfigure}[t]{0.49\linewidth}
        \centering\includegraphics[width=\linewidth]{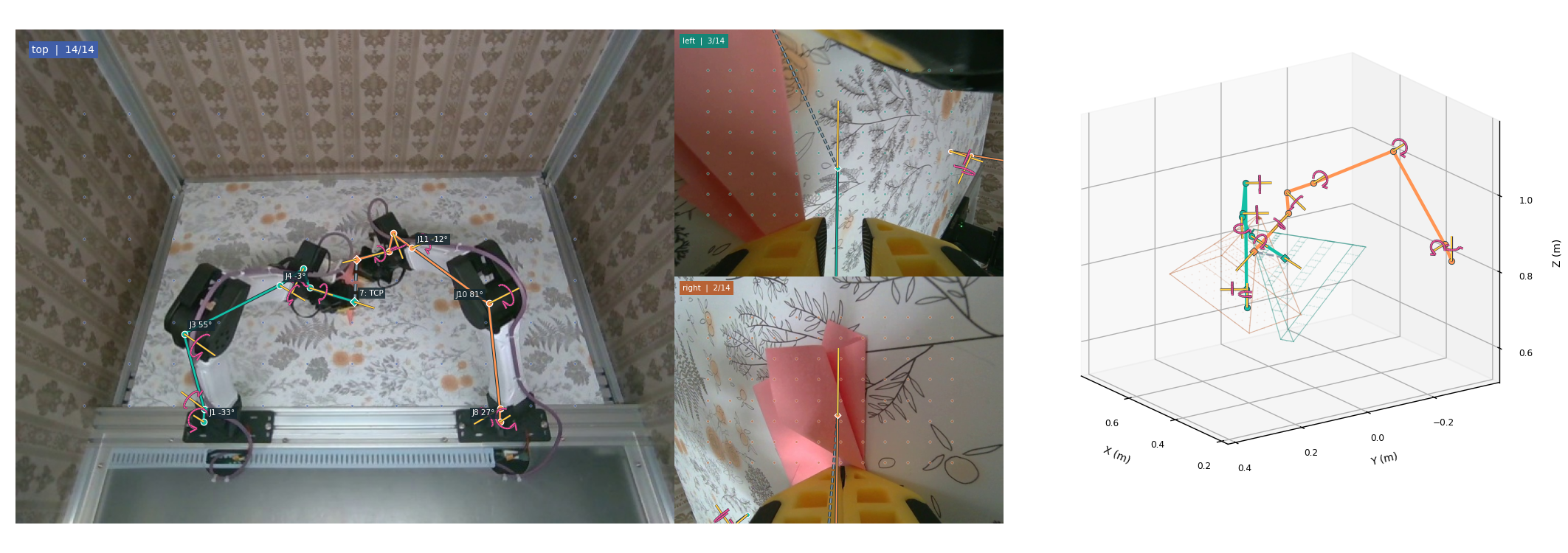}
        \caption{YAM dual-arm robot}
    \end{subfigure}
    \par\medskip
\begin{subfigure}[t]{0.49\linewidth}
        \centering\includegraphics[width=\linewidth]{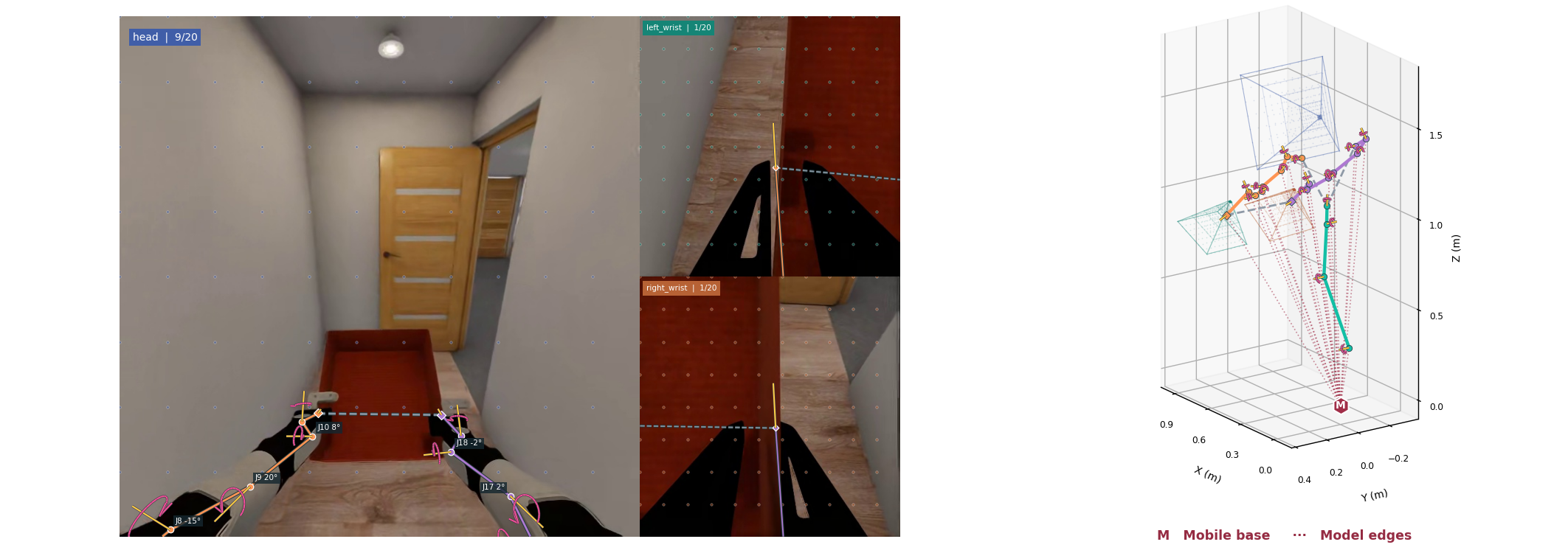}
        \caption{BEHAVIOR R1 Pro mobile base}
    \end{subfigure}\hfill
    \begin{subfigure}[t]{0.49\linewidth}
        \centering\includegraphics[width=\linewidth]{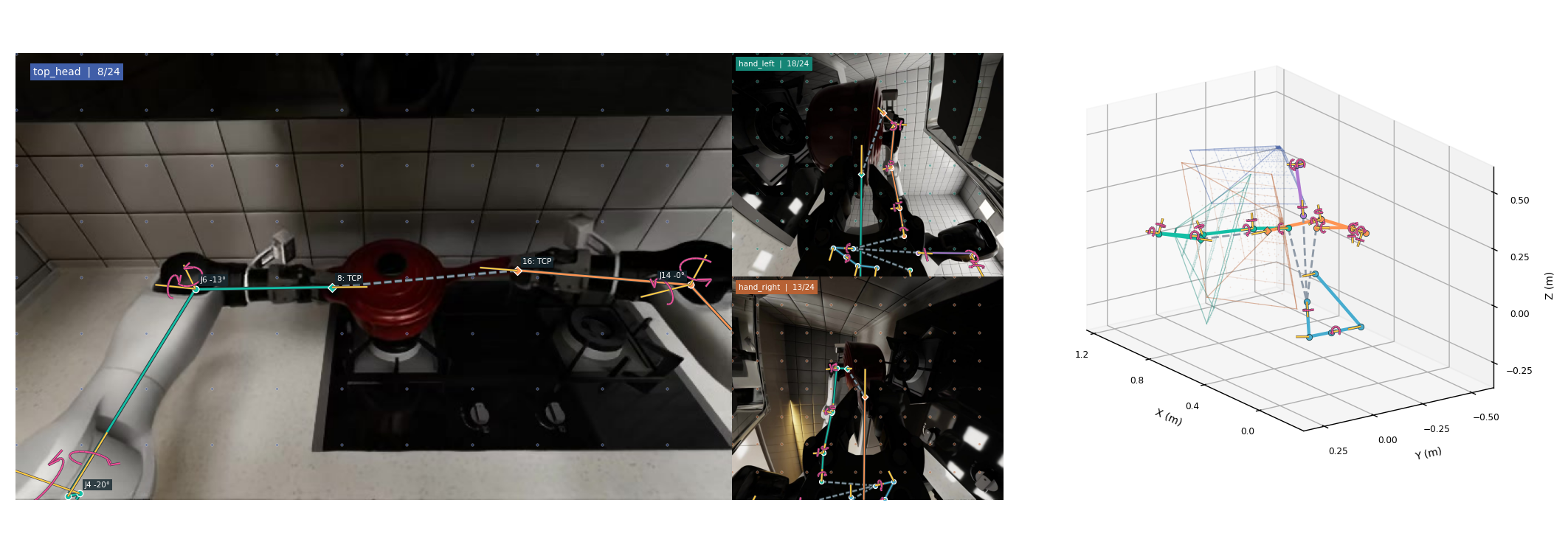}
        \caption{AgiBot G2 OmniPicker}
    \end{subfigure}
    \par\medskip
\begin{subfigure}[t]{0.49\linewidth}
        \centering\includegraphics[width=\linewidth]{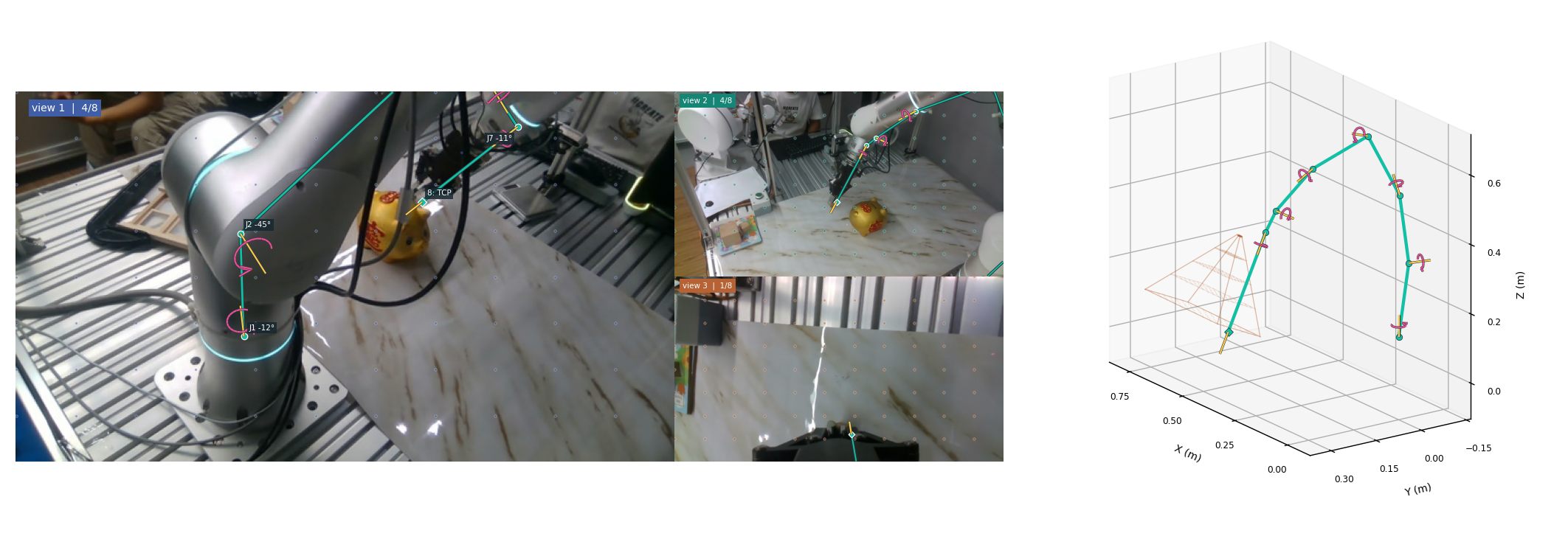}
        \caption{RH20T: Flexiv Rizon}
    \end{subfigure}\hfill
    \begin{subfigure}[t]{0.49\linewidth}
        \centering\includegraphics[width=\linewidth]{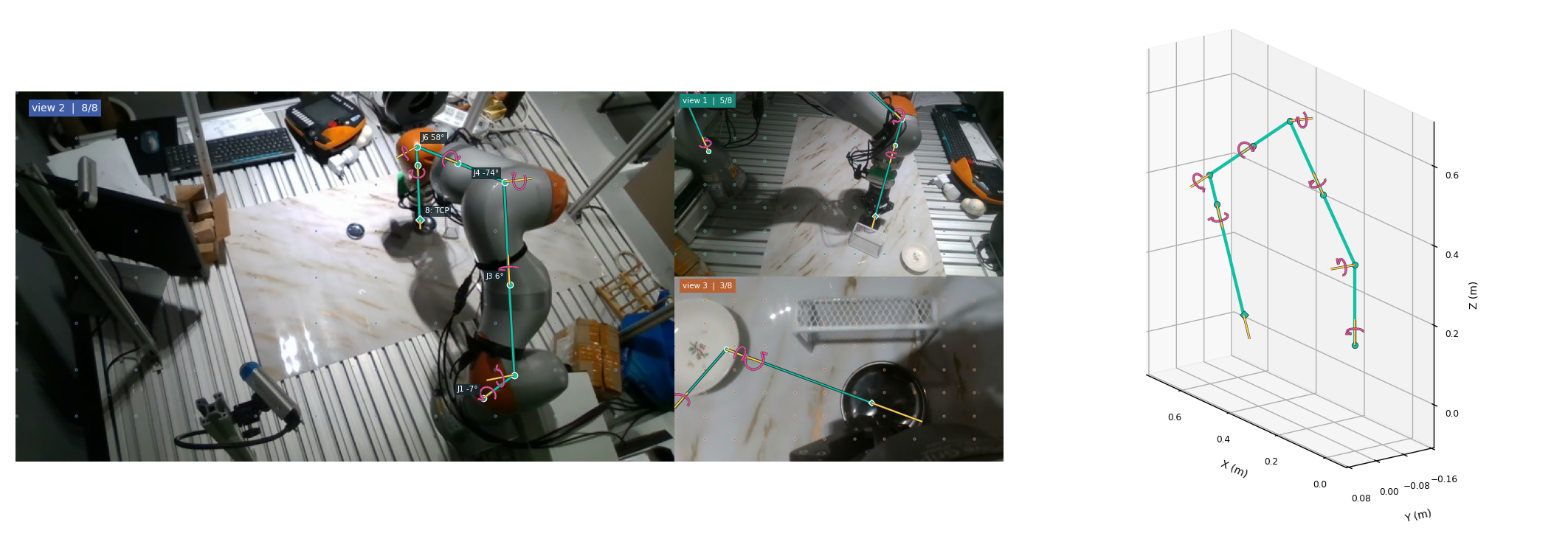}
        \caption{RH20T: KUKA LBR iiwa}
    \end{subfigure}
    \par\medskip
\begin{subfigure}[t]{0.49\linewidth}
        \centering\includegraphics[width=\linewidth]{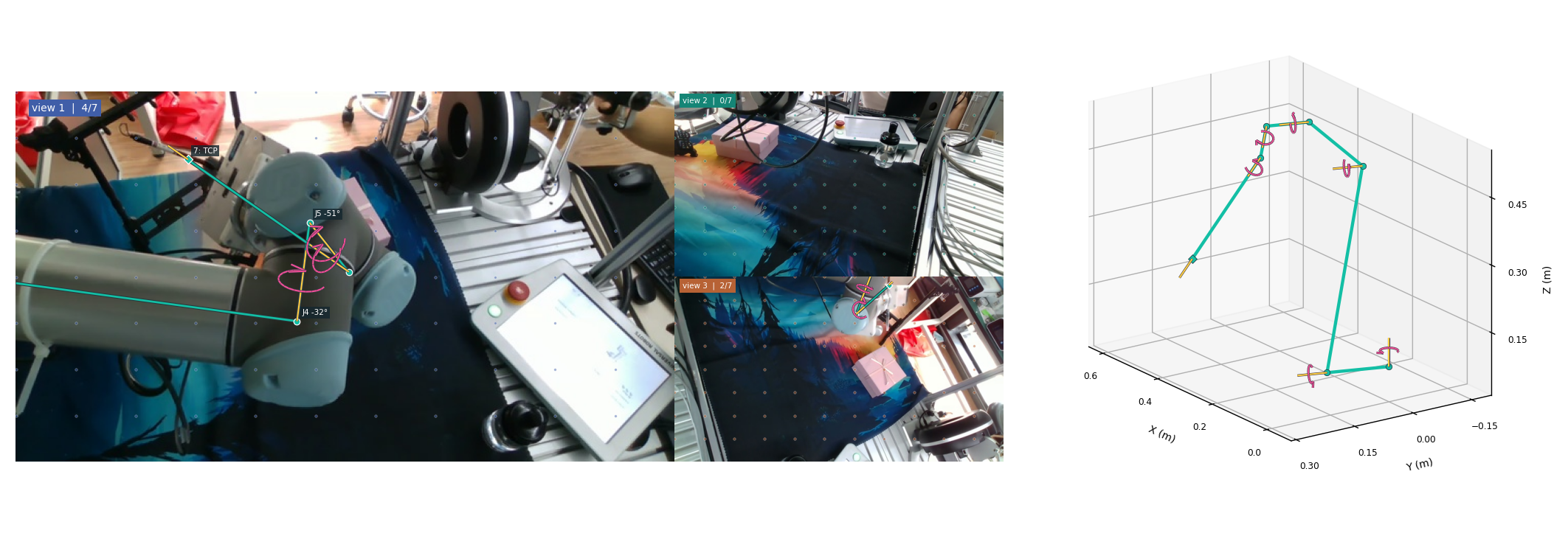}
        \caption{RH20T: UR5}
    \end{subfigure}\hfill
    \begin{subfigure}[t]{0.49\linewidth}
        \centering\includegraphics[width=\linewidth]{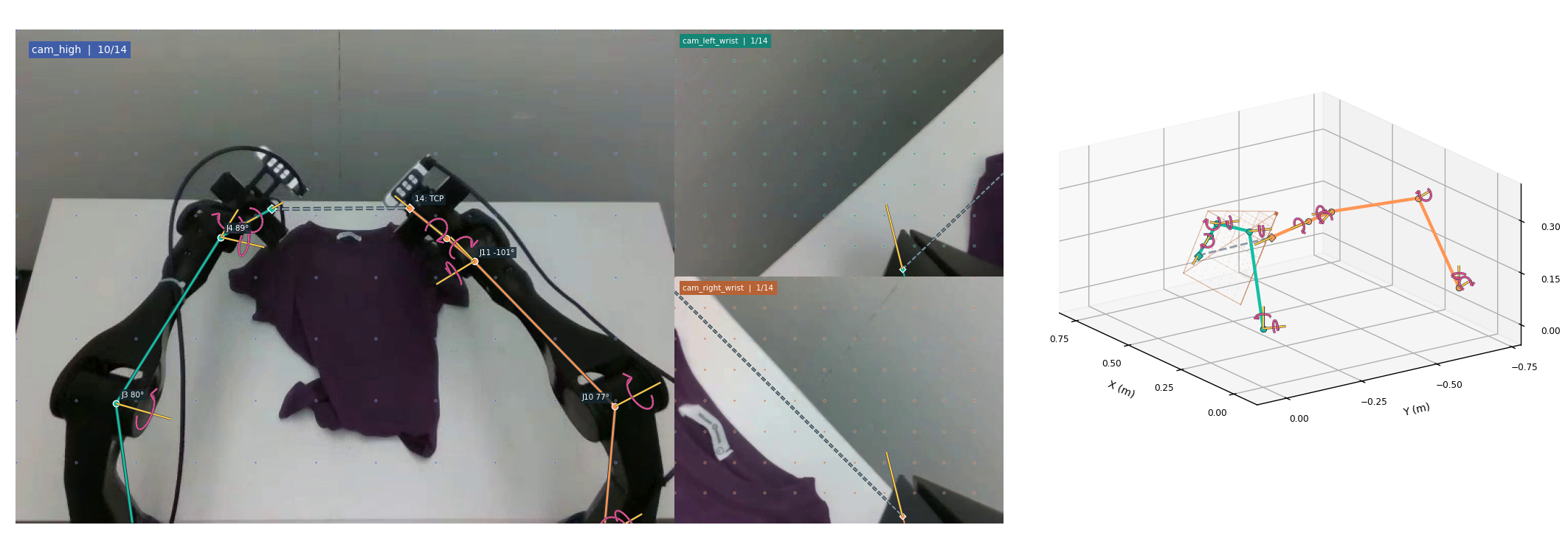}
        \caption{Table30: DOS-W1}
    \end{subfigure}
    \par\medskip
    \begin{subfigure}[t]{0.49\linewidth}
        \centering\includegraphics[width=\linewidth]{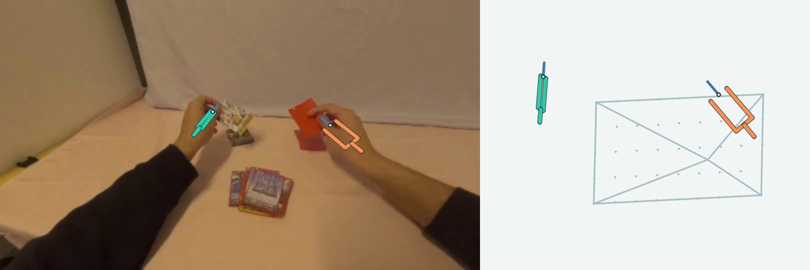}
        \caption{EgoDex: hands represented as grippers}
    \end{subfigure}\hfill
    \begin{subfigure}[t]{0.49\linewidth}
        \centering\includegraphics[width=\linewidth]{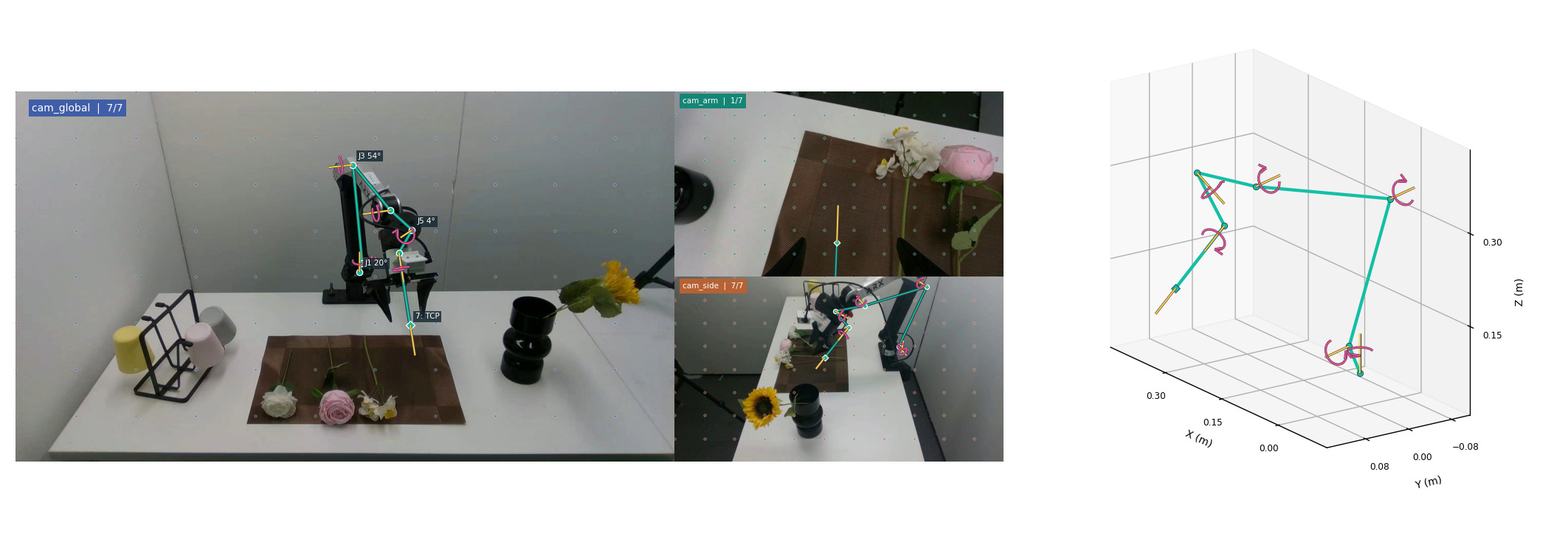}
        \caption{Table30: ARX5}
    \end{subfigure}
    \caption{Additional joint graphs. In each panel, camera views (left) and the 3D graph (right) show the same configuration. The TCP-only interface is represented by one node; the mobile platform also represents motion of the base.}
    \label{fig:additional-graphs}
\end{figure}

\newlength{\vertsep}
\setlength{\vertsep}{.085in}
\newlength{\imsize}
\setlength{\imsize}{.365\textwidth}

\clearpage
\bibliography{paper}
\bibliographystyle{unsrt}

\end{document}